\documentclass[11pt]{article}

\usepackage[final]{acl}

\usepackage{times}
\usepackage{latexsym}

\usepackage[T1]{fontenc}

\usepackage[utf8]{inputenc}
\usepackage{lipsum}%
\usepackage{microtype}

\usepackage{inconsolata}

\usepackage{graphicx}
\usepackage{booktabs}
\usepackage{colortbl}
\usepackage{multirow}
\usepackage{adjustbox}
\usepackage{diagbox}
\usepackage{makecell}
\usepackage{amsmath}
\usepackage{amssymb}

\title{VITAL: Visual-Semantic Dual Supervision for Enhanced and \\ Interpretable Latent Reasoning in Medical MLLMs}

\author{
  \textbf{Qiaoru Li\textsuperscript{1,3}\thanks{Work done during an internship.}},
  \textbf{Shaotian Liang\textsuperscript{1}},
  \textbf{Jintao Chen\textsuperscript{1}},
  \textbf{Haoran Sun\textsuperscript{2}},
  \textbf{Yuxiang Cai\textsuperscript{1,4,5}\thanks{Corresponding authors.}},\\
  \textbf{Jianwei Yin\textsuperscript{1}},
  \textbf{Yankai Jiang\textsuperscript{1,2}\footnotemark[2]}
\\
\\
  \textsuperscript{1}Zhejiang University,
  \textsuperscript{2}Shanghai Artificial Intelligence Laboratory,
  \textsuperscript{3}Tencent,\\
  \textsuperscript{4}Ningbo Global Innovation Center, Zhejiang University,\\
  \textsuperscript{5}Zhejiang Key Laboratory of Digital-Intelligence Service Technology
 \\
  \small{
    \textbf{Correspondence:} \href{mailto:caiyuxiang@zju.edu.cn}{caiyuxiang@zju.edu.cn}, \href{mailto:jiangyankai@pjlab.org.cn}{jiangyankai@pjlab.org.cn}
  }
}

\begin{document}
\maketitle
\begin{abstract}
Latent reasoning enables reasoning over continuous hidden states rather than explicit tokens, avoiding the language bottleneck and inference overhead of chain-of-thought for medical VQA. However, existing methods suffer from modality collapse, insufficient visual supervision, and train--inference mismatch. Moreover, their opaque latent states offer no interpretability, which is critical in clinical applications.
We propose VITAL, a latent-space reasoning framework for medical MLLMs with visual-semantic dual supervision: an auxiliary text decoder reconstructs reasoning chains from latent states, while a visual projector regresses ROI features from a frozen, independent medical vision encoder.
Both modules are discarded at inference with zero overhead, yet can be re-attached post-hoc for dual interpretability, providing textual and visual explanations of the reasoning process without sacrificing efficiency.
We construct a 61K dataset spanning 9 imaging modalities, exceeding prior medical visual latent reasoning datasets by an order of magnitude.
Experiments on 7 benchmarks show that VITAL consistently and substantially outperforms the backbone, all latent reasoning baselines, and medical MLLMs trained on far larger data, achieving state-of-the-art results competitive with trillion-parameter proprietary models. Code is available at \url{https://github.com/EmberQR/vital2026}.
\end{abstract}
\section{Introduction}
\label{sec:intro}

Medical multimodal large language models (MLLMs) have rapidly evolved from image-level question answering toward fine-grained visual understanding \citep{li2023llavamed,chen2024huatuogptvision,xu2025lingshu,shu2025flemingvl}, yet inadequate visual grounding remains a primary bottleneck \citep{liu2026howdomedical}.
This has motivated Chain-of-Thought (CoT) reasoning for medical VQA, decomposing diagnosis into structured multi-step processes \citep{qiao2025medscot,wang2025v2tcot,fan2026stepcot} with expert-annotated visual evidence at each step \citep{le-duc2025schain}.  However, explicit CoT in the multimodal setting suffers from two inherent limitations:
(i)~a language bottleneck: fine-grained visual evidence (lesion texture, boundary morphology, subtle intensity gradients) must be compressed into discrete text tokens, inevitably losing perceptual details critical for clinical judgment;
(ii)~inference overhead: generating verbose reasoning chains substantially increases latency and cost in time-sensitive clinical environments.

\begin{figure*}[t]
\centering
\includegraphics[width=\textwidth]{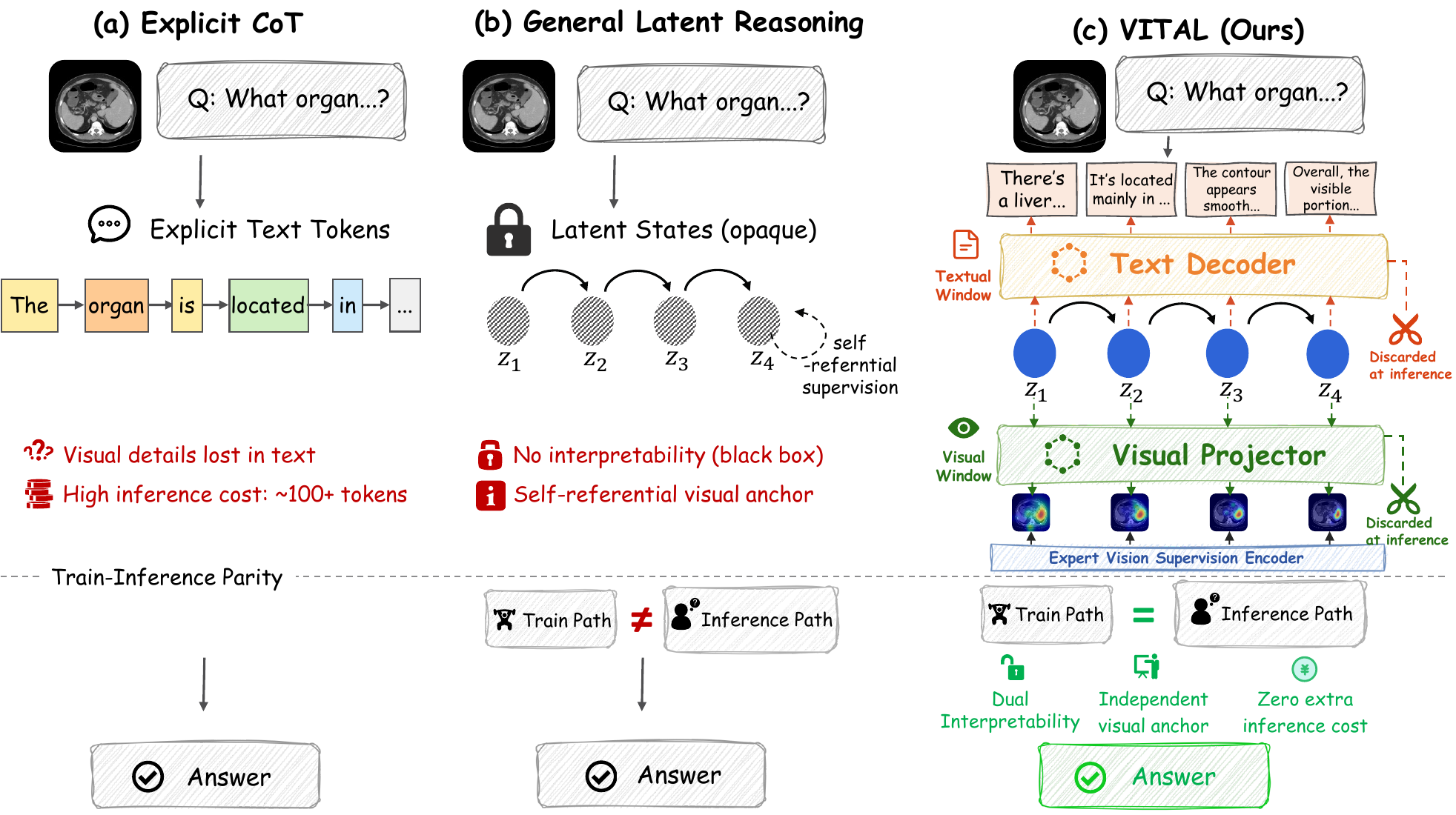}
\caption{\textbf{Comparison of reasoning paradigms.}}
\label{fig:intro_comparison}
\end{figure*}

These limitations motivate latent reasoning, where the model reasons over continuous hidden states rather than explicit text tokens (Figure~\ref{fig:intro_comparison}).
Coconut \citep{hao2024coconut} first demonstrated multi-step latent reasoning by iteratively feeding hidden states back as inputs; subsequent work refined this via self-distillation \citep{shen2025codi}, compact hidden thinking states \citep{shen2025heima}, and step-level supervision to prevent state collapse \citep{wei2025simcot}.
The paradigm has since been extended to vision-language models \citep{li2025latentvisualreasoning,pham2025multimodalchaincontinuousthought}.
In the medical domain, MedLVR \citep{xi2026medlvr} applies latent visual reasoning to medical VQA by aligning latent states with ROI patch tokens from the VLM's own visual encoder and applying RL-based policy optimization.
However, MedLVR exhibits three key limitations:
(i)~its visual supervision is self-referential, aligning to the model's own patch embeddings rather than an independent expert medical vision encoder;
(ii)~its latent states remain black box, offering no post-hoc interpretability;
and (iii)~its two-stage pipeline introduces different computation graphs at training and inference, creating train--test mismatch.

We propose \textbf{VITAL} (\textbf{V}isual-\textbf{I}mplicit \textbf{T}hinking with \textbf{A}ligned \textbf{L}atents), a latent-space reasoning framework for medical MLLMs that addresses all three limitations.
VITAL features a recurrent latent reasoning loop with strict train--inference parity, and introduces visual-semantic dual supervision: an auxiliary text decoder reconstructs the reasoning chain from each latent state, while a visual projector regresses ROI features from a frozen, independent medical vision encoder.
Both auxiliary modules serve as training-time scaffolding, constraining latent states to encode both textual logic and visual evidence. They are \textbf{entirely discarded at inference} with \textbf{zero additional overhead}.
Crucially, these modules can be re-attached post-hoc, yielding dual interpretability: a textual window (what the model reasons) and a visual window (where the model looks), which is a remarkable level of transparency for latent reasoning.

Our contributions are summarized as follows:
\begin{itemize}
\item We propose \textbf{VITAL}, a novel latent reasoning framework with visual-semantic dual supervision for medical MLLMs.
VITAL maintains strict train--inference parity through a single recurrent loop, prevents modality collapse via orthogonal dual supervision anchored to an independent medical vision encoder, and offers dual interpretability, all with zero inference overhead via its plug-and-play scaffolding design.

\item We construct a large-scale dataset of \textbf{$\sim$61K} five-tuple training samples (image, question, answer, $K$-step reasoning chain, ROI visual feature) from two public medical segmentation datasets, spanning \textbf{9 imaging modalities}.
To our knowledge, this is the largest dataset specifically designed for training latent reasoning in the medical vision domain, exceeding prior efforts by $6{\times}$--$57{\times}$ in scale.

\item Extensive experiments on \textbf{7 benchmarks} (2 in-domain testsets, 3 medical VQA benchmarks, 1 visual grounding benchmark, and 1 held-out in-house testset) demonstrate that VITAL consistently improves over the backbone on every evaluation, surpasses all latent reasoning baselines and medical MLLMs trained on far larger data, and achieves performance competitive with trillion-parameter proprietary models while exceeding them on visual grounding by a large margin.
\end{itemize}

\section{Related Work}

\subsection{Multimodal Latent-Space Reasoning}

Coconut \cite{hao2024coconut} pioneered multi-step latent reasoning in LLMs; Heima \cite{shen2025heima} compressed reasoning into compact hidden states with reconstructable traces; and SIM-CoT \cite{wei2025simcot} showed that step-level supervision is essential to prevent latent state collapse.
In the multimodal setting, Latent Visual Reasoning \cite{li2025latentvisualreasoning} interleaves reasoning in visual embedding space with text generation, and MedLVR \cite{xi2026medlvr} first applies this paradigm to medical VQA.
However, MedLVR relies on self-referential visual supervision from the model's own encoder, offers no post-hoc interpretability for latent states, and introduces train--inference mismatch through its two-stage pipeline.

\subsection{Medical Visual Grounding and Multimodal Alignment}

Medical MLLMs have evolved from image-level QA \cite{li2023llavamed,chen2024huatuogptvision} toward fine-grained multimodal reasoning with broader domain coverage \cite{xu2025lingshu,deria2026medmo}.
Meanwhile, grounding capabilities have gained prominence: LISA \cite{lai2024lisa} and GLaMM \cite{rasheed2024glamm} introduced reasoning segmentation and pixel-level grounding in the general domain, with medical extensions (MedPLIB \cite{huang2025medplib}, UniBiomed \cite{wu2025unibiomed}, MMedAgent \cite{li2024mmedagent}, Citrus-V \cite{wang2025citrusv}, and IBISAgent \cite{jiang2026ibisagent}) demonstrating the value of coupling MLLMs with specialized tools and pixel prompts.
However, these methods rely on explicit mask annotations or external tool invocation; direct reasoning in latent visual space for medical understanding remains underexplored, which is the gap our work targets.

\section{Method}

\subsection{Overview of VITAL}
\label{sec:overview}

\begin{figure*}[t]
\centering
\includegraphics[width=\textwidth]{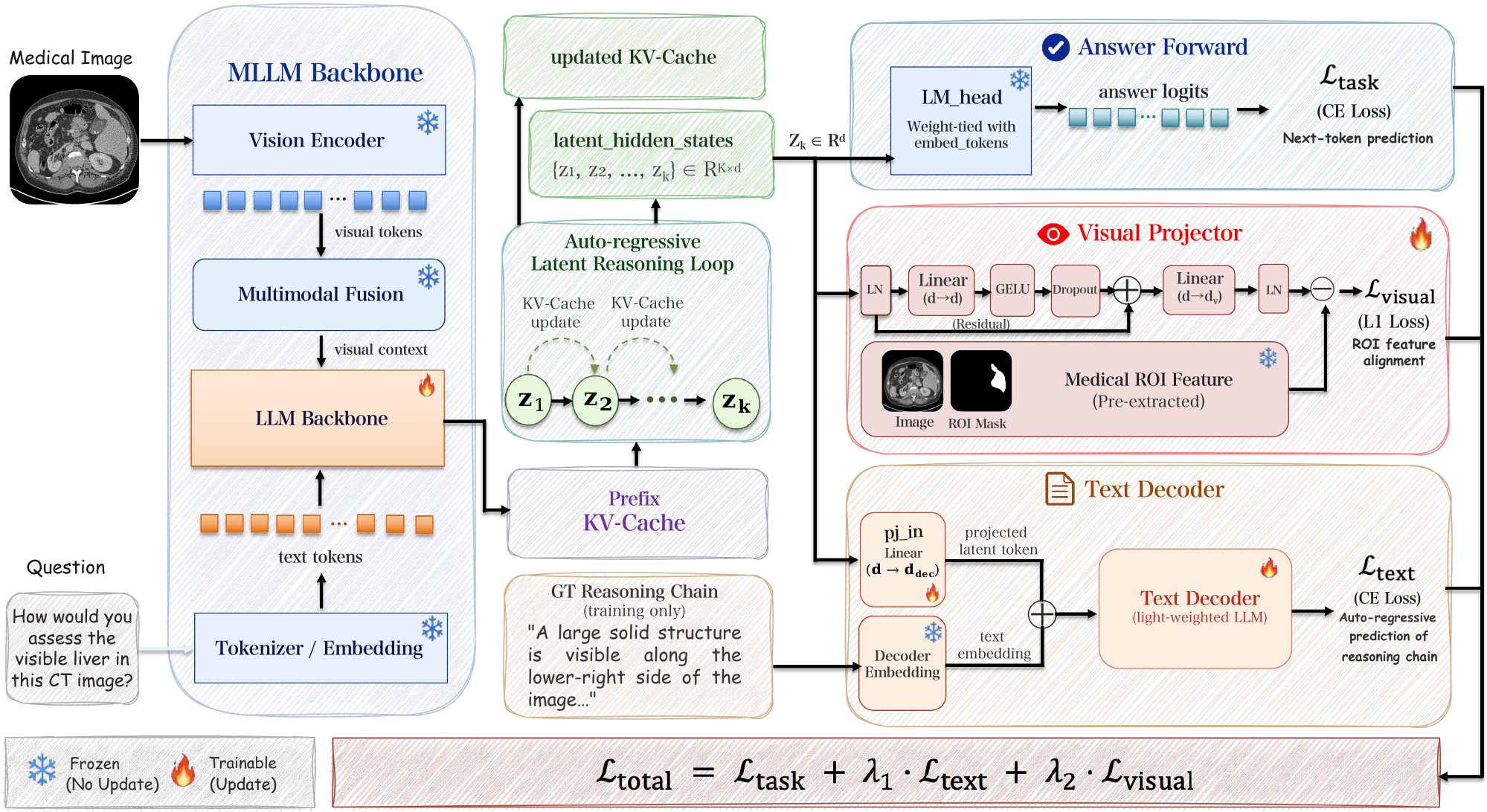}
\caption{\textbf{Overview of VITAL.} The multimodal backbone encodes the input into a prefix KV-cache. A recurrent latent loop iterates $K$ steps ($z_k = f_\theta(z_{k-1}; \mathcal{C})$) with identical paths at training and inference. Latent states are supervised by $\mathcal{L}_{\text{task}}$ (answer), $\mathcal{L}_{\text{text}}$ (auxiliary text decoder), and $\mathcal{L}_{\text{visual}}$ (visual projector regressing ROI features). Both auxiliary modules are discarded at inference with zero overhead.}
\label{fig:framework}

\end{figure*}

VITAL is a latent-space reasoning framework built on a frozen MLLM backbone.
As shown in Figure~\ref{fig:framework}, the forward pass has three stages: (1)~the multimodal backbone encodes the image and question into a prefix KV-cache; (2)~a recurrent latent loop produces $K$ continuous states $\{z_1,\dots,z_K\}$ (\S\ref{sec:latent_loop}); (3)~the answer is auto-regressively decoded from the enriched cache.

Without explicit supervision, latent states may collapse into a text-dominant subspace, discarding fine-grained visual evidence.
VITAL addresses this with visual-semantic dual supervision (\S\ref{sec:dual_supervision}): an auxiliary text decoder and a visual projector constrain latent states from orthogonal directions, yet are \textbf{entirely discarded at inference} (\S\ref{sec:training}).

\subsection{Recurrent Latent Reasoning}
\label{sec:latent_loop}

\paragraph{Prefix encoding.}
Given a medical image $\mathbf{I}$ and a question $\mathbf{q}$, the frozen vision encoder extracts visual tokens which are fused with text tokens through the multimodal backbone.
A full forward pass over the concatenated prefix yields the hidden states and a prefix KV-cache $\mathcal{C}_{\text{prefix}}$.

\paragraph{Recurrent latent loop.}
We adopt a recurrent mechanism to produce $K$ latent reasoning states.
Let $z_0 = h_{\text{prefix}}^{\text{last}} \in \mathbb{R}^{d}$ denote the last hidden state of the prefix.
At each step $k = 1,\dots,K$, the language model performs a single-token forward pass:
\begin{equation}
\label{eq:latent_step}
z_k = f_\theta\!\left(z_{k-1};\; \mathcal{C}_{\text{prefix} \cup \{z_{<k}\}}\right),
\end{equation}
where $f_\theta$ denotes one transformer forward step and the KV-cache is updated in place.
Each step's input is the output of the previous step, creating an auto-regressive causal chain over continuous vectors rather than discrete tokens.

\paragraph{Train--inference parity.}
A critical design is that both training and inference traverse the identical recurrent path: there is no teacher forcing, scheduled sampling, or parallel computation of latent states.
This eliminates train--test mismatch and distinguishes VITAL from prior latent reasoning methods that use different computation graphs at training and inference time.

\paragraph{Answer decoding.}
After the $K$-step loop, the answer token embeddings are fed into the same KV-cache and decoded by the language model head.
Let $\mathbf{a} = (a_1, \dots, a_T)$ denote the answer token sequence of length $T$.
Notably, the first answer token is predicted directly from the final latent state $z_K$, ensuring a smooth transition from continuous reasoning to discrete generation:
\begin{equation}
\label{eq:answer}
p(\mathbf{a}) = \prod_{t=1}^{T} p\!\left(a_t \mid z_{1:K},\, a_{<t}\right).
\end{equation}

\subsection{Visual-Semantic Dual Supervision}
\label{sec:dual_supervision}

Without explicit constraints, latent states in multimodal models tend to collapse into a text-dominant subspace, discarding subtle visual cues such as lesion texture and boundary morphology.
VITAL prevents this modality collapse by supervising every latent state $z_k$ from two directions: semantic and visual (see Figure~\ref{fig:framework}).

\paragraph{Semantic Supervision.}

To ensure that each latent state encodes interpretable reasoning logic, we attach an auxiliary text decoder, an independent, smaller causal language model with its own embedding layer and prediction head (no weight sharing with the backbone).
Let $\mathbf{e}_k$ denote the ground-truth reasoning text for step $k$ in the $K$-step reasoning chain.
For each latent step $k$, the hidden state is first projected by a learnable linear layer $\texttt{pj\_in}: \mathbb{R}^{d} \!\to\! \mathbb{R}^{d_{\text{dec}}}$ to match the decoder's hidden dimension, then concatenated with the decoder's own embeddings of $\mathbf{e}_k$: 
\begin{equation}
\label{eq:text_input}
\mathbf{h}_k^{\text{dec}} = \left[\texttt{pj\_in}(z_k);\; \text{Embed}_{\text{dec}}(\mathbf{e}_k)\right].
\end{equation}
The decoder auto-regressively predicts $\mathbf{e}_k$, and we compute a cross-entropy loss averaged over all $K$ valid steps:
\begin{equation}
\label{eq:text_loss}
\mathcal{L}_{\text{text}} = \frac{1}{K}\sum_{k=1}^{K} \text{CE}\!\left(\text{Decoder}(\mathbf{h}_k^{\text{dec}}),\; \mathbf{e}_k\right).
\end{equation}

\paragraph{Visual Supervision.}

To anchor each latent state to fine-grained visual evidence, we project it into the feature space of a frozen medical vision encoder and regress a pre-extracted ROI feature.
A na\"ive linear projection, however, is prone to training instability due to the severe norm imbalance between the LLM's hidden states and the vision encoder's feature space: \citet{pham2025multimodalchaincontinuousthought} report that the last hidden state norm can be $4{\times}$ larger than the vision feature norm, making direct mapping numerically fragile.

We address this with a visual projector (VP) based on a norm-stabilized residual MLP.
Let $d_v$ denote the feature dimension of the frozen medical vision encoder.
The projector is defined as:
\begin{equation}
\label{eq:vp}
\texttt{VP}(z) = \text{LN}\!\left(W_2\!\left(\text{GELU}\!\left(W_1\,\text{LN}(z)\right) + z\right)\right),
\end{equation}
where $W_1 \!\in\! \mathbb{R}^{d \times d}$ and $W_2 \!\in\! \mathbb{R}^{d_v \times d}$.
The design is motivated by three considerations:
(i)~the \textbf{input LayerNorm} normalizes the LLM hidden state before any linear transformation, eliminating the norm imbalance at source;
(ii)~the \textbf{residual connection} in $d$-dimensional space (around the expansion layer $W_1$ + GELU) preserves the representational richness of $z_k$ while allowing the non-linear pathway to learn a modality-bridging correction;
(iii)~the \textbf{output LayerNorm} re-scales the projected $d_v$-dimensional output to match the $\ell_2$-normalized scale of the vision encoder features, ensuring stable L1 regression.

The supervision target $\mathbf{f}_{\text{ROI}} \!\in\! \mathbb{R}^{d_v}$ is a pre-extracted, $\ell_2$-normalized ROI feature for the target anatomical region (\S\ref{sec:data}).
All $K$ latent steps share the same target, and an L1 loss is applied:
\begin{equation}
\label{eq:visual_loss}
\mathcal{L}_{\text{visual}} = \frac{1}{K}\sum_{k=1}^{K} \left\lVert \texttt{VP}(z_k) - \mathbf{f}_{\text{ROI}} \right\rVert_1.
\end{equation}
We use a shared target for all steps: $\mathcal{L}_{\text{text}}$ already differentiates steps, so the shared visual anchor provides a stable spatial signal without imposing an artificial curriculum (ablation in \S\ref{app:perstep_visual}).

\paragraph{Dual Interpretability.}

As an important by-product of dual supervision, both auxiliary modules can be re-attached during post-hoc analysis. They are discarded at inference time.
The text decoder can decode any latent state into a human-readable reasoning chain, providing a textual window into the model's implicit thought process.
Simultaneously, the visual projector output can be used for nearest-neighbor retrieval in the medical vision encoder's feature space, recovering the visual content that the latent state encodes, which serves as a visual window into what the model ``sees in its mind.''
To the best of our knowledge, this work is among the first to achieve both textual and visual interpretability for latent reasoning.
We provide a qualitative example in Figure~\ref{fig:qua}.

\subsection{Training Data Construction}
\label{sec:data}

VITAL requires each training sample to be a five-tuple: (image, question, answer, $K$-step reasoning chain, ROI visual feature).
We build an automated, reproducible pipeline from two public medical segmentation datasets: MSD~\cite{antonelli2022medical} and BiomedParse~\cite{zhao2025foundation}, producing 61K samples in total.

\paragraph{Teacher distillation.}
Textual reasoning chains and answers are generated by a state-of-the-art proprietary MLLM serving as the teacher.
The teacher receives a teacher-only overlay image in which the segmentation mask is rendered as a semi-transparent color layer on top of the original image, together with the target identity as hidden metadata.
Crucially, the teacher is instructed to produce outputs as if it can only see the raw, unannotated image, leveraging the overlay solely for internal correctness while never leaking annotation cues.
Each question is classified into one of five types (yes/no, identification, location, description, reasoning), and the reasoning depth $K \!\in\! \{1,\dots,4\}$ is set accordingly.
A multi-round quality gate verifies every generated sample against a forbidden-word list to catch annotation leakage, validates reasoning step counts, and normalizes answer format. A subsequent 5\% human audit confirms a leakage rate below 0.3\%.
Full pipeline details, prompt templates, and quality statistics are provided in \S\ref{app:data_pipeline}.

\paragraph{Adaptive ROI feature extraction.}
For each sample, we extract a $d_v$-dimensional ROI feature from a frozen medical vision encoder as the visual supervision target.
To handle the wide range of target sizes (from large organs spanning the full image to small nodules occupying only a few patches), we employ an adaptive extraction strategy: when the mask covers ${\geqslant}20\%$ of the image area, we extract patch features from the full image and pool over mask-aligned patches; when the mask is smaller, we tightly crop the ROI region (with a 5\% margin), pad it to a square, and pool over valid patches.
All features are $\ell_2$-normalized to the unit hypersphere.
See \S\ref{app:roi_extraction} for algorithmic details and threshold ablations.

\subsection{Training Objective and Plug-and-Play Design}
\label{sec:training}

\paragraph{Joint objective.}
The total training loss combines the three supervision signals:
\begin{equation}
\label{eq:total_loss}
\mathcal{L} = \mathcal{L}_{\text{task}} + \lambda_1\,\mathcal{L}_{\text{text}} + \lambda_2\,\mathcal{L}_{\text{visual}},
\end{equation}
where $\mathcal{L}_{\text{task}}$ is the standard next-token cross-entropy loss computed only over answer tokens, and $\lambda_1$, $\lambda_2$ are scalar weights that balance the auxiliary losses.

\paragraph{Curriculum learning.}
Since longer latent chains are harder to optimize from scratch, we adopt a three-phase curriculum:
\textbf{Phase~1} ($K{=}1$) teaches single-step latent reasoning;
\textbf{Phase~2} ($K{\leqslant}2$) introduces two-step chains;
\textbf{Phase~3} ($K{\leqslant}4$) trains on all reasoning depths.
Each phase warm-starts from the previous one.

\paragraph{Plug-and-play inference.}
The auxiliary text decoder, visual projector, and \texttt{pj\_in} serve exclusively as training-time scaffolding.
At inference, all these scaffoldings are discarded: the deployed model consists solely of the frozen backbone augmented with LoRA, the latent loop, and the language model head.
The resulting inference cost is identical to that of a standard LoRA-tuned model with no auxiliary modules: \textbf{zero extra latency, zero extra memory, zero extra parameters}.

\section{Experiment}

\subsection{Experimental Setup}
\label{sec:exp_setup}

\paragraph{Implementation.}
We build VITAL on top of Qwen3-VL-8B~\citep{bai2025qwen3vltechnicalreport}.
The vision encoder and LLM backbone are both frozen; only LoRA adapters ($r{=}64$, $\alpha{=}128$) are inserted into all attention and feed-forward projections of the 36 transformer layers.
The training-time scaffolding consists of an auxiliary text decoder (Qwen3-1.7B, fully fine-tuned), a visual projector, and a projection layer $\pi_{\text{in}}$.
In total, 1.93B out of 10.07B parameters are trainable (19.1\%); at inference all scaffolding is discarded, leaving only the LoRA weights (174.6M).
We train on our 61K five-tuple dataset (\S\ref{sec:data}) with a three-phase curriculum: Phase~1 ($K{=}1$, 5 epochs), Phase~2 ($K{\leqslant}2$, 5 epochs), and Phase~3 (all $K$, 10 epochs), using AdamW with learning rate $2{\times}10^{-5}$, 5\% warmup, and bf16 precision.
All experiments are conducted on 6$\times$NVIDIA RTX PRO 6000 (Blackwell).
Complete hyperparameters and per-module parameter breakdowns are provided in \S\ref{app:impl_details}.

\paragraph{Benchmarks.}
We evaluate on \textbf{7 benchmarks} spanning four categories:
(i)~In-domain testsets: held-out splits of MSD~\citep{antonelli2022medical} and BiomedParse~\citep{zhao2025foundation};
(ii)~Medical VQA: VQA-RAD~\citep{lau2018dataset}, PathVQA~\citep{he2020pathvqa30000questionsmedical}, and PMC-VQA~\citep{zhang2024pmcvqavisualinstructiontuning};
(iii)~Visual grounding: GEMeX-RMCoT~\citep{Liu_2025};
and (iv)~Held-out in-house testset: a private evaluation set of 1{,}000 brain neuroimaging VQA pairs (MRI and PET), curated to avoid training-set contamination from public benchmarks.
For closed-ended questions (yes/no and choice) we report accuracy; for open-ended questions we report token-level F1 (see \S\ref{app:eval_protocol}).

\paragraph{Baselines.}
We compare against four groups of methods (Table~\ref{tab:main_results}).
\textbf{(1)~Text latent reasoning}: Coconut~\citep{hao2024coconut}, CODI~\citep{shen2025codi}, and SIM-CoT~\citep{wei2025simcot}. These methods operate on text-only LLMs and thus have no original multimodal weights; we {\color{red}re-implement} them on our backbone (Qwen3-VL-8B + LoRA) with identical training data, differing only in the reasoning paradigm.
\textbf{(2)~Visual latent reasoning}: MCOUT~\citep{pham2025multimodalchaincontinuousthought} and LVR~\citep{li2025latentvisualreasoning}. We report both their original weights and re-implementations on our backbone.
\textbf{(3)~Medical MLLMs}: LLaVA-Med~\citep{li2023llavamed}, HuatuoGPT-Vision~\citep{chen2024huatuogptvision}, Fleming-VL~\citep{shu2025flemingvl}, and VividMed~\citep{luo2025vividmedvisionlanguagemodel}, evaluated with their official weights.
\textbf{(4)~General MLLMs}: Qwen3-VL-8B (zero-shot), Qwen3-VL-8B-Thinking~\citep{bai2025qwen3vltechnicalreport} (zero-shot explicit CoT from the same backbone family, without fine-tuning on our 61K dataset), GPT-5.4~\citep{gpt54thinking2026}, and Claude-Opus-4.6~\citep{claudeopus46_2026}.

\begin{table*}[t]
\centering
\caption{\textbf{Main results on medical VQA benchmarks.} Text latent reasoning baselines (Coconut, CODI, SIM-CoT) operate on text-only LLMs and thus have no original multimodal weights; we report their {\color{red}re-implementations on our backbone (Qwen3-VL-8B + LoRA) with identical training data}. For visual latent reasoning baselines, we report both original weights and {\color{red}fine-tuned} results. The top three results in each row are in \textbf{bold} and highlighted with an orange background, where the color intensity indicates the rank. The \textsuperscript{\color{red}\textbf{+x.xx}} on VITAL denotes the absolute improvement over Qwen3-VL-8B (zero-shot). All values are percentages (\%).}
\label{tab:main_results}
\adjustbox{max width=\textwidth}{%
\begin{tabular}{l|l|ccc|cc|cccc|cccc|c}
\toprule
\multirow{3}{*}{\diagbox[width=10em, height=3.5em, innerleftsep=3pt, innerrightsep=3pt]{\textbf{Benchmarks}}{\textbf{Methods}}}
& \multirow{3}{*}{\textbf{Metrics}}
& \multicolumn{3}{c|}{\textbf{General Text Latent}}
& \multicolumn{2}{c|}{\textbf{General Visual Latent}}
& \multicolumn{4}{c|}{\textbf{Medical MLLMs}}
& \multicolumn{4}{c|}{\textbf{General MLLMs}}
& \multirow{3}{*}{\textbf{VITAL}} \\
\cmidrule(lr){3-5} \cmidrule(lr){6-7} \cmidrule(lr){8-11} \cmidrule(lr){12-15}
& & \makecell{Coconut} & \makecell{CODI} & \makecell{SIM-CoT}
& \makecell{MCOUT} & \makecell{LVR}
& \makecell{LLaVA-Med} & \makecell{HuatuoGPT-Vision} & \makecell{Fleming-VL} & \makecell{VividMed}
& \makecell{Qwen3-VL-8B \\ (Zero-shot)} & \makecell{Qwen3-VL-8B \\ -Thinking \\ (Zero-shot)} & \makecell{GPT-5.4} & \makecell{Claude-Opus-4.6}
& \\
\midrule

\rowcolor[gray]{0.90}
\multicolumn{16}{l}{\textit{\textbf{In-Domain Testsets}}} \\
\midrule

\multirow{1}{*}{MSD} %
& Acc $\uparrow$     & {\color{red}40.58} & {\color{red}41.09} & {\color{red}47.25} & 27.63\,({\color{red}38.54}) & 30.22\,({\color{red}48.63})  & 46.74 & 55.57 & \cellcolor{orange!40}\textbf{65.50} & 40.19  & 35.56 & 42.52 & \cellcolor{orange!20}\textbf{57.65} & 54.26  & \cellcolor{orange!60}\textbf{83.08}$^{\color{red}\textbf{+47.52}}$ \\
\midrule

\multirow{1}{*}{BiomedParse} %
& Acc $\uparrow$     & {\color{red}38.15} & {\color{red}36.77} & {\color{red}43.37} & 22.88\,({\color{red}38.32}) & 28.85\,({\color{red}44.97})  & 45.31 & 50.27 & \cellcolor{orange!40}\textbf{61.49} & 37.74  & 32.01 & 33.19 & 53.52 & \cellcolor{orange!20}\textbf{55.12}  & \cellcolor{orange!60}\textbf{78.68}$^{\color{red}\textbf{+46.67}}$ \\
\midrule

\rowcolor[gray]{0.90}
\multicolumn{16}{l}{\textit{\textbf{Medical VQA}}} \\
\midrule

\multirow{2}{*}{VQA-RAD} %
& Acc$_\text{closed}$ $\uparrow$ & {\color{red}58.76} & {\color{red}60.31} & {\color{red}64.08} & 35.25\,({\color{red}59.87}) & 39.47\,({\color{red}63.19})  & 58.96 & 70.29 & 69.62 & 45.45  & 67.63 & 64.52 & \cellcolor{orange!60}\textbf{81.67} & \cellcolor{orange!40}\textbf{77.29}  & \cellcolor{orange!20}\textbf{74.28}$^{\color{red}\textbf{+6.65}}$ \\
& Token-F1$_\text{open}$ $\uparrow$   & {\color{red}33.23} & {\color{red}28.14} & {\color{red}25.93} & 5.62\,({\color{red}30.19}) & 3.29\,({\color{red}27.76})  & 27.45 & \cellcolor{orange!40}\textbf{34.35} & \cellcolor{orange!60}\textbf{37.94} & 7.29  & 31.50 & 26.77 & 16.73 & 3.07  & \cellcolor{orange!20}\textbf{32.59}$^{\color{red}\textbf{+1.09}}$ \\
\midrule

\multirow{2}{*}{PathVQA} %
& Acc$_\text{closed}$ $\uparrow$ & {\color{red}56.14} & {\color{red}56.05} & {\color{red}62.57} & 41.14\,({\color{red}58.52}) & 45.23\,({\color{red}64.70})  & 57.56 & 65.38 & 64.10 & 46.90  & 60.80 & 64.86 & \cellcolor{orange!20}\textbf{72.06} & \cellcolor{orange!40}\textbf{72.85}  & \cellcolor{orange!60}\textbf{73.95}$^{\color{red}\textbf{+13.15}}$ \\
& Token-F1$_\text{open}$ $\uparrow$   & {\color{red}1.18} & {\color{red}2.09} & {\color{red}5.32} & 1.23\,({\color{red}5.14}) & 1.41\,({\color{red}4.69})  & 2.32 & 5.06 & \cellcolor{orange!60}\textbf{25.82} & 1.94  & 7.25 & \cellcolor{orange!20}\textbf{16.33} & 3.51 & 1.62  & \cellcolor{orange!40}\textbf{18.95}$^{\color{red}\textbf{+11.70}}$ \\
\midrule

\multirow{1}{*}{PMC-VQA} %
& Acc $\uparrow$     & {\color{red}50.20} & {\color{red}48.60} & {\color{red}51.30} & 25.60\,({\color{red}48.89}) & 30.40\,({\color{red}51.56})  & 13.85 & 53.20 & 61.30 & 31.15  & 54.65 & 55.75 & \cellcolor{orange!60}\textbf{65.80} & \cellcolor{orange!40}\textbf{64.80}  & \cellcolor{orange!20}\textbf{61.70}$^{\color{red}\textbf{+7.05}}$ \\
\midrule

\rowcolor[gray]{0.90} %
\multicolumn{16}{l}{\textit{\textbf{Visual Grounding QA}}} \\
\midrule

\multirow{1}{*}{GEMeX-RMCoT}
& Acc $\uparrow$     & {\color{red}41.67} & {\color{red}38.33} & {\color{red}40.00} & 26.67\,({\color{red}45.00}) & 21.67\,({\color{red}38.33})  & 43.33 & 41.67 & 40.00 & 33.33  & 36.67 & 41.67 & \cellcolor{orange!20}\textbf{56.67} & \cellcolor{orange!40}\textbf{58.33}  & \cellcolor{orange!60}\textbf{86.67}$^{\color{red}\textbf{+50.00}}$ \\
\midrule

\rowcolor[gray]{0.90} %
\multicolumn{16}{l}{\textit{\textbf{Held-out In-house Testset}}} \\
\midrule

\diagbox[width=10em, height=1em]{}{}
& Acc $\uparrow$     & {\color{red}41.50} & {\color{red}48.40} & {\color{red}52.90} & 38.20\,({\color{red}51.50}) & 39.70\,({\color{red}52.70})  & 45.70 & 66.10 & \cellcolor{orange!20}\textbf{73.60} & 47.10  & 52.60 & 55.30 & \cellcolor{orange!40}\textbf{73.80} & 72.50  & \cellcolor{orange!60}\textbf{80.50}$^{\color{red}\textbf{+27.90}}$ \\
\bottomrule
\end{tabular}%
}
\end{table*}

\subsection{Main Results}
\paragraph{VQA Performance.}
Table~\ref{tab:main_results} presents comprehensive results across 7 benchmarks.
A striking finding is that all re-implemented general latent reasoning methods (Coconut, CODI, SIM-CoT, MCOUT, LVR) degrade the Qwen3-VL-8B backbone after fine-tuning on the same data (e.g., Coconut scores only 58.76 on VQA-RAD closed-ended, below the zero-shot backbone's 67.63), indicating that na\"ive latent reasoning can harm multimodal performance.
In contrast, VITAL consistently improves over the zero-shot backbone on every single benchmark, demonstrating that our visual-semantic dual supervision effectively prevents modality collapse during latent training.
Notably, VITAL trained on only 61K samples outperforms medical MLLMs that are trained on datasets orders of magnitude larger, highlighting the data efficiency of well-supervised latent reasoning.
Compared with Qwen3-VL-8B-Thinking (a zero-shot explicit-CoT model from the same backbone family), VITAL achieves substantially higher accuracy across all benchmarks (e.g., +40.56 on MSD, +45.49 on BiomedParse, +45.00 on GEMeX-RMCoT) while being ${\sim}$97$\times$ faster at inference (see~\S\ref{par:ablation_efficiency}), confirming that latent reasoning in a well-supervised continuous space is far more effective than verbose token-level reasoning chains for medical VQA.
On closed-ended metrics, VITAL surpasses all latent reasoning baselines and all medical MLLMs, improving over Qwen3-VL-8B by +47.52 on MSD, +46.67 on BiomedParse, +13.15 on PathVQA, and +7.05 on PMC-VQA; the only methods that outperform VITAL on individual splits are trillion-parameter proprietary models.
On visual grounding (GEMeX-RMCoT), VITAL achieves 86.67, exceeding the best proprietary model by 28.34, strongly validating that latent states encode fine-grained spatial evidence.
On the held-out in-house testset, VITAL reaches 80.50, outperforming GPT-5.4 (73.80) and all medical MLLMs, confirming robust generalization beyond the training distribution.
On open-ended Token-F1, VITAL trails medical MLLMs optimized for verbose free-text generation (e.g., Fleming-VL), reflecting an inherent limitation of latent reasoning: it produces concise answers that score lower on token-overlap metrics designed to reward verbose outputs.

\paragraph{Comparison with Matched Verbalized-CoT Fine-Tuning.}
The Qwen3-VL-8B-Thinking baseline above evaluates zero-shot explicit CoT and is not fine-tuned on our data.
To separately isolate the contribution of our latent reasoning framework from that of the training data, we train a matched verbalized-CoT baseline using the same Qwen3-VL-8B backbone, LoRA configuration, and 61K samples as VITAL, but generating the reasoning chain as explicit text tokens.
VITAL outperforms this baseline by 13.21 points on the in-domain testsets, 11.75 points on VQA-RAD, and 26.40 points on the held-out in-house testset, while being $58.4\times$ faster (353\,ms vs.\ 20.6\,s).
These controlled results demonstrate that VITAL's gains arise from its latent reasoning framework with dual supervision rather than from data exposure alone.
Full results are provided in \S\ref{app:verbalized_cot} and Table~\ref{tab:verbalized_cot}.

\begin{figure}[t]
\centering
\includegraphics[width=\columnwidth]{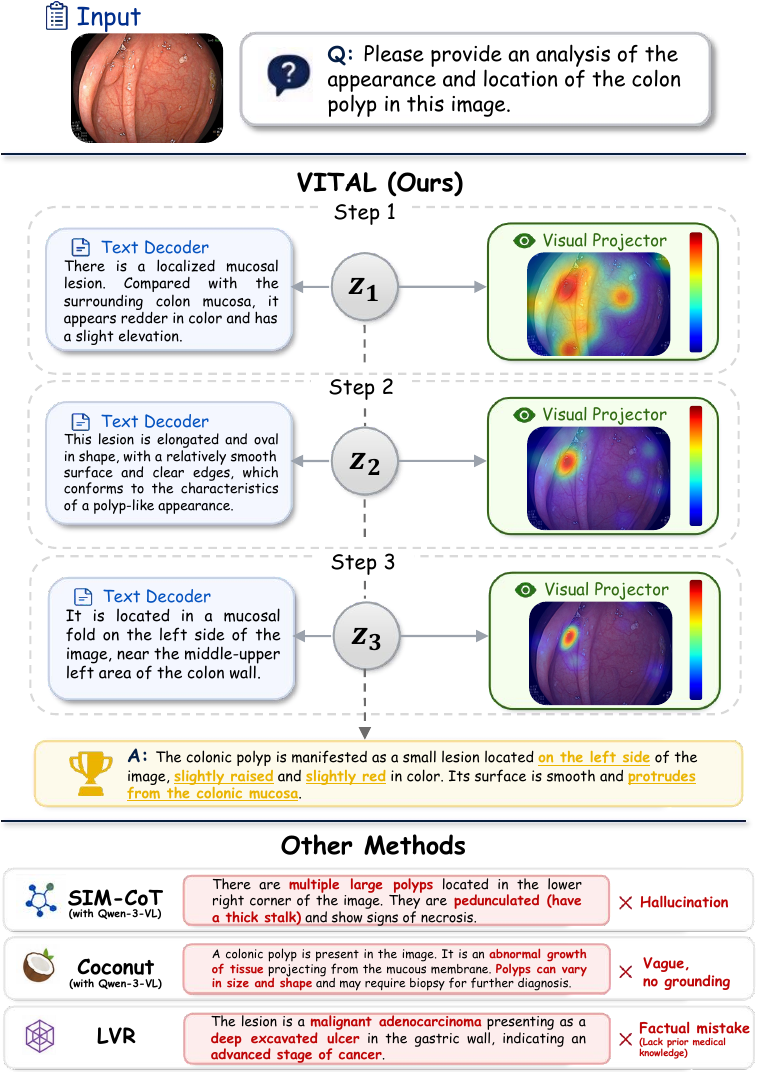}
\caption{\textbf{Qualitative comparison.} VITAL achieves precise visual grounding and accurate diagnosis via progressive latent reasoning ($Z_1 \to Z_3$). Conversely, baselines struggle with hallucinations, ungrounded vagueness, or medical factual errors.}
\label{fig:qua}
\end{figure}

\paragraph{Qualitative Analysis.}
Figure~\ref{fig:qua} illustrates VITAL's dual interpretability on a colonoscopy case asking about polyp appearance and location.
Through progressive latent reasoning ($z_1 \!\to\! z_3$), the text decoder reveals an increasingly refined chain: $z_1$ identifies a localized mucosal lesion with reddish color and slight elevation; $z_2$ characterizes its elongated, smooth-surfaced, polyp-like morphology; and $z_3$ pinpoints its position on the left mucosal fold.
Simultaneously, the visual projector heatmaps progressively sharpen from a broad activation over the mucosal region to a tight focus on the polyp itself, confirming that latent states encode spatially grounded evidence.
The final answer correctly synthesizes all four aspects (location, color, surface texture, and protrusion).
In contrast, SIM-CoT hallucinates ``multiple large polyps'' with fabricated morphology in a wrong location; Coconut produces a vague, ungrounded description without any spatial specificity; and LVR commits a severe factual error by diagnosing ``malignant adenocarcinoma'' in the ``gastric wall'', misidentifying both the pathology and the organ.
These failures highlight that, without visual-semantic dual supervision, latent reasoning can collapse into text-dominant hallucination or lose medical factual grounding.

\paragraph{Quantitative Interpretability.}
Beyond the qualitative cases in Figure~\ref{fig:qua}, we quantitatively evaluate whether the re-attached auxiliary modules faithfully recover information from VITAL's latent states.
On 9{,}582 valid step-level pairs, the text decoder achieves an overall BERTScore-F1 of 85.14 and ROUGE-L of 52.02, indicating that the semantic content of the reasoning process remains recoverable.
For visual interpretation, the final latent state achieves 74.2 pointing accuracy and 51.2 patch IoU against ground-truth masks, with localization improving consistently across reasoning steps.
These results provide quantitative support for VITAL's textual and visual interpretability. Full metric definitions and step-wise results are reported in \S\ref{app:auxiliary_reliability}.

\subsection{Ablation Studies}

\paragraph{Effect of dual supervision.}
\label{par:ablation_dual}
We ablate the two auxiliary supervision signals by removing them individually and jointly.
Table~\ref{tab:ablation_dual} reports average accuracy on three representative evaluation groups: in-domain testsets, VQA-RAD (closed-ended), and the held-out in-house testset. Task-only latent training (b) improves over the untuned backbone (a) on in-domain and in-house sets but drops on VQA-RAD ($-$9.32), indicating that unsupervised latent states can overfit to the training distribution. Adding either supervision alone restores and extends gains: $\mathcal{L}_\text{visual}$ (d) contributes more on in-domain (+22.32 over b) and VQA-RAD (+13.31), while $\mathcal{L}_\text{text}$ (c) provides moderate but consistent improvement across all sets. Crucially, the full dual supervision (e) outperforms both single-signal variants by a large margin, confirming that the two losses are complementary rather than redundant: semantic supervision prevents reasoning degradation while visual supervision anchors latent states to spatial evidence. Ablation on $\lambda_1, \lambda_2$ sensitivity are in~\S\ref{app:loss_weight}.

\begin{table}[t]
\centering
\small
\caption{\textbf{Ablation on dual supervision components.} Avg.\ accuracy (\%) is reported.}
\label{tab:ablation_dual}
\setlength{\tabcolsep}{3.5pt}
\begin{tabular}{@{}lcccccc@{}}
\toprule
\textbf{Variant} & $\mathcal{L}_\text{task}$ & $\mathcal{L}_\text{text}$ & $\mathcal{L}_\text{visual}$ & \textbf{In-D.} & \textbf{RAD} & \textbf{In-H.} \\
\midrule
(a) Backbone     &            &            &            & 33.97 & 67.63 & 52.60 \\
(b) Task-Only    & \checkmark &            &            & 45.47 & 58.31 & 56.60 \\
(c) +Semantic    & \checkmark & \checkmark &            & 52.29 & 62.08 & 64.10 \\
(d) +Visual      & \checkmark &            & \checkmark & 67.79 & 71.62 & 70.60 \\
\rowcolor{orange!15}
(e) \textbf{VITAL}        & \checkmark & \checkmark & \checkmark & \textbf{81.08} & \textbf{74.28} & \textbf{80.50} \\
\bottomrule
\end{tabular}
\end{table}

\paragraph{Effect of reasoning depth $K$.}
We vary the number of latent reasoning steps $K$ from 0 (no latent) to 4. As shown in~Table~\ref{tab:ablation_K}, $K{=}0$ substantially underperforms all latent variants, confirming that the recurrent latent loop is essential.
Performance improves monotonically with $K$: the largest single-step gain occurs from $K{=}1$ to $K{=}2$ (+16.04 on in-domain, +10.10 on in-house), after which returns diminish: $K{=}3$ to $K{=}4$ gains only +4.05/+0.22/+0.80 across the three sets.
We use $K{=}4$ as the default since it achieves the best accuracy with still-acceptable latency (see~\S\ref{par:ablation_efficiency}).
\begin{table}[t]
\centering
\footnotesize
\caption{\textbf{Ablation on the number of latent steps $K$ and curriculum learning strategy.} Avg.\ accuracy (\%) is reported.}
\label{tab:ablation_K}
\setlength{\tabcolsep}{3.5pt}
\begin{minipage}[t]{0.44\columnwidth}
\centering
\begin{tabular}{@{}cccc@{}}
\toprule
$K$ & \textbf{In-D.} & \textbf{RAD} & \textbf{In-H.} \\
\midrule
0 & 41.52 & 69.62 & 63.80 \\
1 & 58.67 & 71.40 & 68.50 \\
2 & 74.71 & 72.95 & 78.60 \\
3 & 77.03 & 74.06 & 79.70 \\
\rowcolor{orange!15}
4 & \textbf{81.08} & \textbf{74.28} & \textbf{80.50} \\
\bottomrule
\end{tabular}
\end{minipage}%
\hfill
\begin{minipage}[t]{0.54\columnwidth}
\centering
\begin{tabular}{@{}lccc@{}}
\toprule
\textbf{Strategy} & \textbf{In-D.} & \textbf{RAD} & \textbf{In-H.} \\
\midrule
(a) Full-mix  & 44.82 & 65.14 & 50.70 \\
(b) 2-Phase   & 56.15 & 68.33 & 64.10 \\
\rowcolor{orange!15}
(c) 3-Phase   & \textbf{81.08} & \textbf{74.28} & \textbf{80.50} \\
(d) Reverse   & 35.23 & 47.15 & 41.40 \\
\bottomrule
\end{tabular}
\end{minipage}
\label{tab:ablation_curriculum}
\end{table}

\begin{figure*}[t]
\centering
\includegraphics[width=\textwidth]{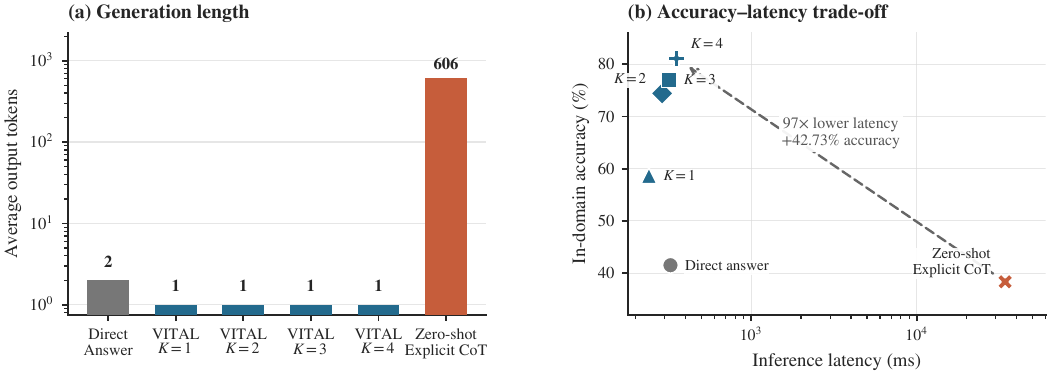}
\caption{\textbf{Inference efficiency.} (a) Average output tokens on a logarithmic scale. (b) In-domain accuracy vs.\ inference latency on a logarithmic scale. ``Zero-shot Explicit CoT'' denotes Qwen3-VL-8B-Thinking without fine-tuning on our 61K dataset. VITAL achieves substantially higher accuracy with concise outputs; at $K{=}4$, it is ${\sim}$97$\times$ faster than zero-shot Explicit CoT.}
\label{fig:efficiency}
\end{figure*}

\paragraph{Inference efficiency.}
\label{par:ablation_efficiency}
Figure~\ref{fig:efficiency} compares VITAL against Direct Answer ($K{=}0$) and Qwen3-VL-8B-Thinking, a zero-shot explicit-CoT model from the same backbone family that is labeled ``Explicit CoT'' in the figure.
This zero-shot Thinking-CoT baseline generates verbose reasoning chains, incurring 34.1s latency, which is \textbf{97$\times$} slower than VITAL ($K{=}4$, 353ms), yet achieves only 38.35 accuracy, far below even the no-latent baseline.
In contrast, VITAL at $K{=}4$ reaches 81.08 accuracy with latency comparable to direct answer (353ms vs.\ 325ms), demonstrating that latent reasoning delivers substantial quality gains at negligible cost.
Notably, VITAL at $K{=}1$ is actually faster than direct answer (241ms) due to shorter output sequences, while already improving accuracy by +17.15.

\paragraph{Effect of curriculum strategy.}
We compare four training schedules in Table~\ref{tab:ablation_curriculum}: (a)~full-mix trains on all $K{\in}\{1,2,3,4\}$ from the start; (b)~2-phase splits into $K{\leqslant}2$ then all $K$; (c)~3-phase (our default) progressively introduces $K{=}1$, $K{\leqslant}2$, then all $K$; (d)~reverse starts from $K{=}4$ and decreases. Full-mix training struggles to optimize long chains from scratch, while the reverse curriculum performs worst overall (35.23 on in-domain), confirming that an easy-to-hard progression is critical for stable latent state learning.
The 3-phase schedule outperforms 2-phase by +24.93/+5.95/+16.40, indicating that isolating single-step reasoning in Phase~1 builds a stronger foundation before introducing multi-step chains.
More ablation studies are presented in~\S\ref{app:more_experiments}.

\section{Conclusion}

We presented VITAL, a latent-space reasoning framework for medical MLLMs that introduces visual-semantic dual supervision to prevent modality collapse while maintaining strict train--inference parity and zero inference overhead.
The auxiliary text decoder and visual projector serve as training-time scaffolding that can be re-attached post-hoc for dual interpretability, revealing both the reasoning logic and spatial attention of latent states.
VITAL consistently outperforms all latent reasoning baselines and medical MLLMs on 7 benchmarks, achieving performance competitive with trillion-parameter proprietary models and surpassing them on visual grounding.
We hope VITAL demonstrates that effective latent reasoning in multimodal models requires explicit grounding in both semantic and visual spaces, opening a path toward efficient, interpretable clinical AI.
Along this path, we envision models that learn when to stop thinking, adaptively allocating reasoning depth to question complexity, and that draw supervision from richer narrative sources, bridging the gap between pixel-level evidence and clinical language at scale.

\section{Acknowledgements}
This work is supported by the Zhejiang Key Laboratory Project (2024E10001).

\section{Limitations}

\paragraph{Data scale and modality coverage.}
Although our 61K five-tuple dataset is the largest medical visual latent reasoning dataset to date, it covers only 9 imaging modalities drawn from two public segmentation datasets (MSD and BiomedParse).
Generalization to rare modalities (e.g., ultrasound elastography, OCT) and uncommon diseases has not been fully validated.

\paragraph{Dependence on teacher-generated reasoning chains.}
The training reasoning chains are distilled from a proprietary MLLM teacher.
The quality ceiling of VITAL's latent states is therefore bounded by the teacher's medical reasoning ability, and potential errors or biases in the teacher may propagate through distillation.

\paragraph{Fixed reasoning depth.}
The current design uses a fixed maximum number of latent steps $K{=}4$ regardless of question difficulty.
Simple questions may not require multi-step reasoning, while more complex cases could benefit from additional steps.
Exploring adaptive early-exit mechanisms is a promising direction for future work.

\paragraph{Open-ended generation.}
Experiments show that VITAL's open-ended Token-F1 trails medical MLLMs specifically optimized for verbose free-text generation.
This reflects an inherent trade-off of the latent reasoning paradigm, which prioritizes answer correctness over output verbosity.

\section{Ethical Considerations}

\paragraph{Not intended for clinical deployment.}
VITAL is a research contribution and should not be used directly for clinical diagnostic decisions.
Any real-world application would require rigorous clinical validation, regulatory approval, and deployment under physician supervision.

\paragraph{Data and privacy.}
All training data are derived from publicly released, de-identified medical datasets (MSD and BiomedParse).
No patient-identifiable information is used during teacher distillation or model training.

\paragraph{Limitations of interpretability outputs.}
While VITAL provides dual interpretability through re-attachable auxiliary modules, the decoded reasoning chains and visual projector outputs are approximate visualizations of latent states rather than faithful representations of the model's true decision process.
Users should not treat these explanations as the sole basis for clinical judgment, nor should they over-rely on the completeness of the interpretability outputs.

\bibliography{custom}

@article{antonelli2022medical,
  title={The medical segmentation decathlon},
  author={Antonelli, Michela and Reinke, Annika and Bakas, Spyridon and Farahani, Keyvan and Kopp-Schneider, Annette and Landman, Bennett A and Litjens, Geert and Menze, Bjoern and Ronneberger, Olaf and Summers, Ronald M and others},
  journal={Nature communications},
  volume={13},
  number={1},
  pages={4128},
  year={2022},
  publisher={Nature Publishing Group UK London}
}

@article{zhao2025foundation,
  title={A foundation model for joint segmentation, detection and recognition of biomedical objects across nine modalities},
  author={Zhao, Theodore and Gu, Yu and Yang, Jianwei and Usuyama, Naoto and Lee, Ho Hin and Kiblawi, Sid and Naumann, Tristan and Gao, Jianfeng and Crabtree, Angela and Abel, Jacob and others},
  journal={Nature methods},
  volume={22},
  number={1},
  pages={166--176},
  year={2025},
  publisher={Nature Publishing Group US New York}
}

@inproceedings{gai2025medthink,
    title = "{M}ed{T}hink: A Rationale-Guided Framework for Explaining Medical Visual Question Answering",
    author = "Gai, Xiaotang  and
      Zhou, Chenyi  and
      Liu, Jiaxiang  and
      Feng, Yang  and
      Wu, Jian  and
      Liu, Zuozhu",
    editor = "Chiruzzo, Luis  and
      Ritter, Alan  and
      Wang, Lu",
    booktitle = "Findings of the Association for Computational Linguistics: NAACL 2025",
    month = apr,
    year = "2025",
    address = "Albuquerque, New Mexico",
    publisher = "Association for Computational Linguistics",
    url = "https://aclanthology.org/2025.findings-naacl.415/",
    doi = "10.18653/v1/2025.findings-naacl.415",
    pages = "7453--7465",
    ISBN = "979-8-89176-195-7"
}

@article{jiang2026m3cotbench,
  title={M3CoTBench: Benchmark Chain-of-Thought of MLLMs in Medical Image Understanding},
  author={Jiang, Juntao and Zhang, Jiangning and Bi, Yali and Bai, Jinsheng and Liu, Weixuan and Jin, Weiwei and Xue, Zhucun and Liu, Yong and Hu, Xiaobin and Yan, Shuicheng},
  journal={arXiv preprint arXiv:2601.08758},
  year={2026}
}

@techreport{gemini31pro2026,
  author      = {{Google DeepMind}},
  title       = {Gemini 3.1 Pro Model Card},
  institution = {Google DeepMind},
  year        = {2026},
  url         = {https://deepmind.google/models/model-cards/gemini-3-1-pro/}
}

@techreport{gpt54thinking2026,
    author      = {{OpenAI}},
    title       = {GPT-5.4 Thinking System Card},
    institution = {OpenAI},
    year        = {2026},
    url         = {https://deploymentsafety.openai.com/gpt-5-4-thinking/gpt-5-4-thinking.pdf}
}

@misc{claudeopus46_2026,
    author = {{Anthropic}},
    title  = {Introducing Claude Opus 4.6},
    year   = {2026},
    url    = {https://www.anthropic.com/news/claude-opus-4-6}
}

@article{zhang2023biomedclip,
  title={Biomedclip: a multimodal biomedical foundation model pretrained from fifteen million scientific image-text pairs},
  author={Zhang, Sheng and Xu, Yanbo and Usuyama, Naoto and Xu, Hanwen and Bagga, Jaspreet and Tinn, Robert and Preston, Sam and Rao, Rajesh and Wei, Mu and Valluri, Naveen and others},
  journal={arXiv preprint arXiv:2303.00915},
  year={2023}
}

@article{sellergren2025medgemma,
  title={Medgemma technical report},
  author={Sellergren, Andrew and Kazemzadeh, Sahar and Jaroensri, Tiam and Kiraly, Atilla and Traverse, Madeleine and Kohlberger, Timo and Xu, Shawn and Jamil, Fayaz and Hughes, C{\'\i}an and Lau, Charles and others},
  journal={arXiv preprint arXiv:2507.05201},
  year={2025}
}

@article{xi2026medlvr,
  title={MedLVR: Latent Visual Reasoning for Reliable Medical Visual Question Answering},
  author={Xi, Suyang and Hu, Songtao and Lai, Yuxiang and Dan, Wangyun and Liu, Yaqi and Wang, Shansong and Yang, Xiaofeng},
  journal={arXiv preprint arXiv:2604.09757},
  year={2026}
}

@article{pham2025multimodalchaincontinuousthought,
  title={Multimodal chain of continuous thought for latent-space reasoning in vision-language models},
  author={Pham, Tan-Hanh and Ngo, Chris},
  journal={arXiv preprint arXiv:2508.12587},
  year={2025}
}

@article{hao2024coconut,
  title={Training large language models to reason in a continuous latent space},
  author={Hao, Shibo and Sukhbaatar, Sainbayar and Su, DiJia and Li, Xian and Hu, Zhiting and Weston, Jason and Tian, Yuandong},
  journal={arXiv preprint arXiv:2412.06769},
  year={2024}
}

@article{shen2025heima,
  title={Efficient reasoning with hidden thinking},
  author={Shen, Xuan and Wang, Yizhou and Zhou, Yufa and Shi, Xiangxi and Zhao, Pu and Wang, Yanzhi and Gu, Jiuxiang},
  journal={arXiv preprint arXiv:2501.19201},
  year={2025}
}

@misc{li2025latentvisualreasoning,
      title={Latent Visual Reasoning}, 
      author={Bangzheng Li and Ximeng Sun and Jiang Liu and Ze Wang and Jialian Wu and Xiaodong Yu and Hao Chen and Emad Barsoum and Muhao Chen and Zicheng Liu},
      year={2025},
      eprint={2509.24251},
      archivePrefix={arXiv},
      primaryClass={cs.CV},
      url={https://arxiv.org/abs/2509.24251}, 
}

@misc{wei2025simcot,
      title={SIM-CoT: Supervised Implicit Chain-of-Thought}, 
      author={Xilin Wei and Xiaoran Liu and Yuhang Zang and Xiaoyi Dong and Yuhang Cao and Jiaqi Wang and Xipeng Qiu and Dahua Lin},
      year={2025},
      eprint={2509.20317},
      archivePrefix={arXiv},
      primaryClass={cs.CL},
      url={https://arxiv.org/abs/2509.20317}, 
}

@inproceedings{li2023llavamed,
 author = {Li, Chunyuan and Wong, Cliff and Zhang, Sheng and Usuyama, Naoto and Liu, Haotian and Yang, Jianwei and Naumann, Tristan and Poon, Hoifung and Gao, Jianfeng},
 booktitle = {Advances in Neural Information Processing Systems},
 editor = {A. Oh and T. Naumann and A. Globerson and K. Saenko and M. Hardt and S. Levine},
 pages = {28541--28564},
 publisher = {Curran Associates, Inc.},
 title = {LLaVA-Med: Training a Large Language-and-Vision Assistant for Biomedicine in One Day},
 url = {https://proceedings.neurips.cc/paper_files/paper/2023/file/5abcdf8ecdcacba028c6662789194572-Paper-Datasets_and_Benchmarks.pdf},
 volume = {36},
 year = {2023}
}

@misc{chen2024huatuogptvision,
      title={HuatuoGPT-Vision, Towards Injecting Medical Visual Knowledge into Multimodal LLMs at Scale}, 
      author={Junying Chen and Chi Gui and Ruyi Ouyang and Anningzhe Gao and Shunian Chen and Guiming Hardy Chen and Xidong Wang and Ruifei Zhang and Zhenyang Cai and Ke Ji and Guangjun Yu and Xiang Wan and Benyou Wang},
      year={2024},
      eprint={2406.19280},
      archivePrefix={arXiv},
      primaryClass={cs.CV},
      url={https://arxiv.org/abs/2406.19280}, 
}

@misc{xu2025lingshu,
  title={Lingshu: A generalist foundation model for unified multimodal medical understanding and reasoning},
  author={Xu, Weiwen and Chan, Hou Pong and Li, Long and Aljunied, Mahani and Yuan, Ruifeng and Wang, Jianyu and Xiao, Chenghao and Chen, Guizhen and Liu, Chaoqun and Li, Zhaodonghui and others},
  journal={arXiv preprint arXiv:2506.07044},
  year={2025}
}

@misc{deria2026medmo,
  title={MedMO: Grounding and Understanding Multimodal Large Language Model for Medical Images},
  author={Deria, Ankan and Kumar, Komal and Dukre, Adinath Madhavrao and Segal, Eran and Khan, Salman and Razzak, Imran},
  journal={arXiv preprint arXiv:2602.06965},
  year={2026}
}

@InProceedings{lai2024lisa,
    author    = {Lai, Xin and Tian, Zhuotao and Chen, Yukang and Li, Yanwei and Yuan, Yuhui and Liu, Shu and Jia, Jiaya},
    title     = {LISA: Reasoning Segmentation via Large Language Model},
    booktitle = {Proceedings of the IEEE/CVF Conference on Computer Vision and Pattern Recognition (CVPR)},
    month     = {June},
    year      = {2024},
    pages     = {9579-9589}
}

@InProceedings{rasheed2024glamm,
    author    = {Rasheed, Hanoona and Maaz, Muhammad and Shaji, Sahal and Shaker, Abdelrahman and Khan, Salman and Cholakkal, Hisham and Anwer, Rao M. and Xing, Eric and Yang, Ming-Hsuan and Khan, Fahad S.},
    title     = {GLaMM: Pixel Grounding Large Multimodal Model},
    booktitle = {Proceedings of the IEEE/CVF Conference on Computer Vision and Pattern Recognition (CVPR)},
    month     = {June},
    year      = {2024},
    pages     = {13009-13018}
}

@inproceedings{huang2025medplib,
  title={Towards a multimodal large language model with pixel-level insight for biomedicine},
  author={Huang, Xiaoshuang and Shen, Lingdong and Liu, Jia and Shang, Fangxin and Li, Hongxiang and Huang, Haifeng and Yang, Yehui},
  booktitle={Proceedings of the AAAI Conference on Artificial Intelligence},
  volume={39},
  number={4},
  pages={3779--3787},
  year={2025}
}

@misc{wu2025unibiomed,
  title={Unibiomed: A universal foundation model for grounded biomedical image interpretation},
  author={Wu, Linshan and Nie, Yuxiang and He, Sunan and Zhuang, Jiaxin and Luo, Luyang and Li, Tao and Xie, Zhuoyao and Chen, Dexuan and Zhao, Yinghua and Mahboobani, Neeraj and others},
  journal={arXiv preprint arXiv:2504.21336},
  year={2025}
}

@misc{li2024mmedagent,
  title={Mmedagent: Learning to use medical tools with multi-modal agent},
  author={Li, Binxu and Yan, Tiankai and Pan, Yuanting and Luo, Jie and Ji, Ruiyang and Ding, Jiayuan and Xu, Zhe and Liu, Shilong and Dong, Haoyu and Lin, Zihao and others},
  booktitle={Findings of the Association for Computational Linguistics: EMNLP 2024},
  pages={8745--8760},
  year={2024}
}

@misc{wang2025citrusv,
  title={Citrus-V: Advancing Medical Foundation Models with Unified Medical Image Grounding for Clinical Reasoning},
  author={Wang, Guoxin and Zhao, Jun and Liu, Xinyi and Liu, Yanbo and Cao, Xuyang and Li, Chao and Liu, Zhuoyun and Sun, Qintian and Zhou, Fangru and Xing, Haoqiang and others},
  journal={arXiv preprint arXiv:2509.19090},
  year={2025}
}

@misc{jiang2026ibisagent,
  title={IBISAgent: Reinforcing Pixel-Level Visual Reasoning in MLLMs for Universal Biomedical Object Referring and Segmentation},
  author={Jiang, Yankai and Li, Qiaoru and Xu, Binlu and Sun, Haoran and Ding, Chao and Dong, Junting and Cai, Yuxiang and Zhang, Xuhong and Yin, Jianwei},
  journal={arXiv preprint arXiv:2601.03054},
  year={2026}
}

@inproceedings{shen2025codi,
  title={Codi: Compressing chain-of-thought into continuous space via self-distillation},
  author={Shen, Zhenyi and Yan, Hanqi and Zhang, Linhai and Hu, Zhanghao and Du, Yali and He, Yulan},
  booktitle={Proceedings of the 2025 Conference on Empirical Methods in Natural Language Processing},
  pages={677--693},
  year={2025}
}

@inproceedings{liu2026howdomedical,
  title={How Do Medical MLLMs Fail? A Study on Visual Grounding in Medical Images},
  author={Liu, Guimeng and Yu, Tianze and Ebrahimkhani, Somayeh and Shawn, Lin Zhi Zheng and Ng, Kok Pin and Cheung, Ngai-Man},
  journal={arXiv preprint arXiv:2603.14323},
  year={2026}
}

@article{le-duc2025schain,
  title={S-chain: Structured visual chain-of-thought for medicine},
  author={Le-Duc, Khai and Nguyen, Duy MH and Trinh, Phuong TH and Nguyen, Tien-Phat and Diep, Nghiem T and Ngo, An and Vu, Tung and Vuong, Trinh and Nguyen, Anh-Tien and Nguyen, Mau and others},
  journal={arXiv preprint arXiv:2510.22728},
  year={2025}
}

@inproceedings{wang2025v2tcot,
  title={V2t-cot: From vision to text chain-of-thought for medical reasoning and diagnosis},
  author={Wang, Yuan and Liu, Jiaxiang and Gao, Shujian and Feng, Bin and Tang, Zhihang and Gai, Xiaotang and Wu, Jian and Liu, Zuozhu},
  booktitle={International Conference on Medical Image Computing and Computer-Assisted Intervention},
  pages={658--668},
  year={2025},
  organization={Springer}
}

@misc{fan2026stepcot,
  title={Step-CoT: Stepwise Visual Chain-of-Thought for Medical Visual Question Answering},
  author={Fan, Lin and Ou, Yafei and Deng, Zhipeng and Dai, Pengyu and Chongxian, Hou and Yan, Jiale and Li, Yaqian and Long, Kaiwen and Gong, Xun and Ikebe, Masayuki and others},
  journal={arXiv preprint arXiv:2603.13878},
  year={2026}
}

@article{qiao2025medscot,
  title={Med-SCoT: Structured chain-of-thought reasoning and evaluation for enhancing interpretability in medical visual question answering},
  author={Qiao, Jinhao and Li, Sihan and Liu, Jiang and Yu, Heng and Xiao, Yi and Yu, Hongshan and Zheng, Yan},
  journal={Computerized Medical Imaging and Graphics},
  pages={102659},
  year={2025},
  publisher={Elsevier}
}

@article{shu2025flemingvl,
  title={Fleming-VL: Towards Universal Medical Visual Reasoning with Multimodal LLMs},
  author={Shu, Yan and Liu, Chi and Chen, Robin and Li, Derek and Dai, Bryan},
  journal={arXiv preprint arXiv:2511.00916},
  year={2025}
}

@misc{luo2025vividmedvisionlanguagemodel,
  title={Vividmed: Vision language model with versatile visual grounding for medicine},
  author={Luo, Lingxiao and Tang, Bingda and Chen, Xuanzhong and Han, Rong and Chen, Ting},
  booktitle={Proceedings of the 2025 Conference of the Nations of the Americas Chapter of the Association for Computational Linguistics: Human Language Technologies (Volume 1: Long Papers)},
  pages={1800--1821},
  year={2025}
}

@misc{bai2025qwen3vltechnicalreport,
  title={Qwen3-vl technical report},
  author={Bai, Shuai and Cai, Yuxuan and Chen, Ruizhe and Chen, Keqin and Chen, Xionghui and Cheng, Zesen and Deng, Lianghao and Ding, Wei and Gao, Chang and Ge, Chunjiang and others},
  journal={arXiv preprint arXiv:2511.21631},
  year={2025}
}

@article{lau2018dataset,
  title={A dataset of clinically generated visual questions and answers about radiology images},
  author={Lau, Jason J and Gayen, Soumya and Ben Abacha, Asma and Demner-Fushman, Dina},
  journal={Scientific data},
  volume={5},
  number={1},
  pages={180251},
  year={2018},
  publisher={Nature Publishing Group}
}

@misc{he2020pathvqa30000questionsmedical,
  title={Pathvqa: 30000+ questions for medical visual question answering},
  author={He, Xuehai and Zhang, Yichen and Mou, Luntian and Xing, Eric and Xie, Pengtao},
  journal={arXiv preprint arXiv:2003.10286},
  year={2020}
}

@misc{zhang2024pmcvqavisualinstructiontuning,
  title={Pmc-vqa: Visual instruction tuning for medical visual question answering, 2024},
  author={Zhang, Xiaoman and Wu, Chaoyi and Zhao, Ziheng and Lin, Weixiong and Zhang, Ya and Wang, Yanfeng and Xie, Weidi},
  journal={URL https://arxiv. org/abs/2305.10415},
  volume={40},
  year={2024} 
}

@inproceedings{Liu_2025,
  title={GEMeX-RMCoT: An Enhanced Med-VQA Dataset for Region-Aware Multimodal Chain-of-Thought Reasoning},
  author={Liu, Bo and Zhao, Xiangyu and He, Along and Chen, Yidi and Fu, Huazhu and Wu, Xiao-Ming},
  booktitle={Proceedings of the 33rd ACM International Conference on Multimedia},
  pages={13213--13220},
  year={2025}
}

\appendix

\section{Training Data Construction}
\label{app:data_pipeline}

VITAL requires each training sample to be a five-tuple $(\mathbf{I}, \mathbf{q}, \mathbf{a}, \{\mathbf{e}_k\}_{k=1}^K, \mathbf{f}_{\text{ROI}})$: a medical image, a question, the final answer, a $K$-step reasoning chain, and a pre-extracted ROI visual feature.
We construct an automated, fully reproducible pipeline from two public medical segmentation datasets, producing approximately 61K high-quality samples in total.
This appendix describes each stage in detail.

\subsection{Source Datasets and Preprocessing}
\label{app:source_data}

\paragraph{MSD.}
We use the Medical Segmentation Decathlon (MSD)~\citep{antonelli2022medical} as our primary source of normal-anatomy data.
MSD provides 3D CT/MRI volumes with voxel-level segmentation annotations for 10 organ or tumor classes: liver, lung, pancreas, hepatic vessels, spleen, colon, heart, brain tumor, prostate, and hippocampus.
For each volume, we extract 2D axial slices and retain only those containing at least 100 pixels of the target organ mask, yielding high-quality image--mask pairs.
Since MSD annotations cover only the target regions without pathological findings, we apply an additional pathology-term filter during teacher distillation (\S\ref{app:quality_control}) to prevent hallucinated clinical descriptions.

\paragraph{BiomedParse.}
To cover pathological findings and diverse biomedical structures beyond normal anatomy, we incorporate the BiomedParse dataset~\citep{zhao2025foundation}, a large-scale multimodal biomedical segmentation corpus spanning multiple target types: organ, lesion, tumor, vessel, tissue, cell, and retinal structure.
Unlike MSD, BiomedParse samples frequently contain abnormal findings, allowing us to train the model on clinically relevant reasoning such as lesion characterization and differential description.

\paragraph{From segmentation to QA tuples.}
Starting from the raw segmentation annotations, we first construct a four-tuple for each sample: $(\texttt{image\_path}, \texttt{mask}, \texttt{target\_name}, \texttt{target\_type})$.
Questions are then auto-generated using a template-based strategy conditioned on the target type and imaging modality.
We define six question categories, namely \emph{yes/no}, \emph{identify}, \emph{location-choice}, \emph{location}, \emph{describe}, and \emph{reasoning}, each associated with a target reasoning depth $K \in \{1, \dots, 4\}$ (Table~\ref{tab:question_types}).
The result is a preliminary dataset containing images, masks, user questions with placeholder answers, and metadata (question type, target name, target type, imaging modality).

\begin{table}[h]
\centering
\caption{\textbf{Question types and associated reasoning depth.}}
\label{tab:question_types}
\small
\begin{tabular}{@{}lcp{3.8cm}@{}}
\toprule
\textbf{Type} & \textbf{$K$} & \textbf{Example Trigger} \\
\midrule
Yes/No & 1 & Is the liver visible in... \\
Identify & 1 & What organ is shown... \\
Loc.-Choice & 1 & Is X in left, right, or center... \\
Location & 2 & Where is X located... \\
Describe & 2--3 & Describe the appearance of... \\
Reasoning & 3--4 & Analyze the visual findings... \\
\bottomrule
\end{tabular}
\end{table}

\subsection{Teacher Distillation via Proprietary MLLM}
\label{app:teacher_distillation}

We employ a state-of-the-art proprietary MLLM as the teacher model to generate high-quality reasoning chains and final answers for each sample.

\paragraph{Teacher model selection.}
The quality of distilled reasoning chains critically depends on the teacher's medical visual reasoning capability.
To select the most suitable teacher, we conduct a pilot study on a random subset of 2{,}000 samples from our preliminary dataset, evaluating three frontier MLLMs (Gemini-3.1-Pro-Preview~\cite{gemini31pro2026}, GPT-5.4~\cite{gpt54thinking2026}, and Claude-Opus-4.6~\cite{claudeopus46_2026}) under identical prompting conditions (same system prompt, overlay images, and quality filters).
We measure: (i) \emph{reasoning accuracy}, defined as whether the final answer correctly identifies/locates the target, judged by a medical annotator; (ii) \emph{leakage rate}, i.e., the fraction of outputs containing forbidden annotation-related terms; and (iii) \emph{format compliance}, i.e., the fraction of outputs that are valid JSON with correct step counts on the first attempt.

\begin{table}[h]
\centering
\caption{\textbf{Teacher model comparison on 2{,}000-sample pilot set.}}
\label{tab:teacher_selection}
\small
\resizebox{\columnwidth}{!}{%
\begin{tabular}{@{}lccc@{}}
\toprule
\textbf{Model} & \textbf{Acc.\,(\%)} $\uparrow$ & \textbf{Leak.\,(\%)} $\downarrow$ & \textbf{Format\,(\%)} $\uparrow$ \\
\midrule
Gemini-3.1-Pro-Preview & 37.85 & 35.10 & 44.10 \\
GPT-5.4 & \textbf{89.35} & \textbf{9.45} & \textbf{97.20} \\
Claude-Opus-4.6 & 66.60 & 12.15 & 78.50 \\
\bottomrule
\end{tabular}%
}
\end{table}

As shown in Table~\ref{tab:teacher_selection}, GPT-5.4 achieves the best overall performance across all three metrics, combining the highest reasoning accuracy with the lowest annotation leakage rate and near-perfect format compliance.
We therefore adopt GPT-5.4 as our teacher model for the full-scale distillation pipeline.

\paragraph{Teacher-only overlay.}
A key challenge in teacher distillation is to generate reasoning chains that are both \emph{spatially grounded} and \emph{annotation-free}. 
If the teacher only observes the raw medical image, it may fail to localize the exact target region, especially for small lesions, subtle tissues, or ambiguous anatomical structures. 
In contrast, if the teacher explicitly refers to the segmentation mask or annotated region, the generated reasoning may leak supervision cues and become unsuitable for training a student model that only receives raw images.

To address this issue, we adopt a \emph{teacher-only overlay} mechanism. 
For each training sample, the binary segmentation mask is rendered on top of the original image as a semi-transparent red overlay with transparency $\alpha=0.4$. 
In our implementation, the overlay color is set to RGB $(255,0,0)$, and the mask boundary is additionally highlighted with a yellow contour, RGB $(255,255,0)$, to help the teacher localize the target region more reliably. 
The resulting overlay image is shown \emph{only} to the teacher model during reasoning-chain distillation, while the student model is trained and evaluated using the original unannotated image.

\begin{figure}[t]
\centering
\begin{minipage}[t]{0.48\linewidth}
\centering
\includegraphics[width=\linewidth]{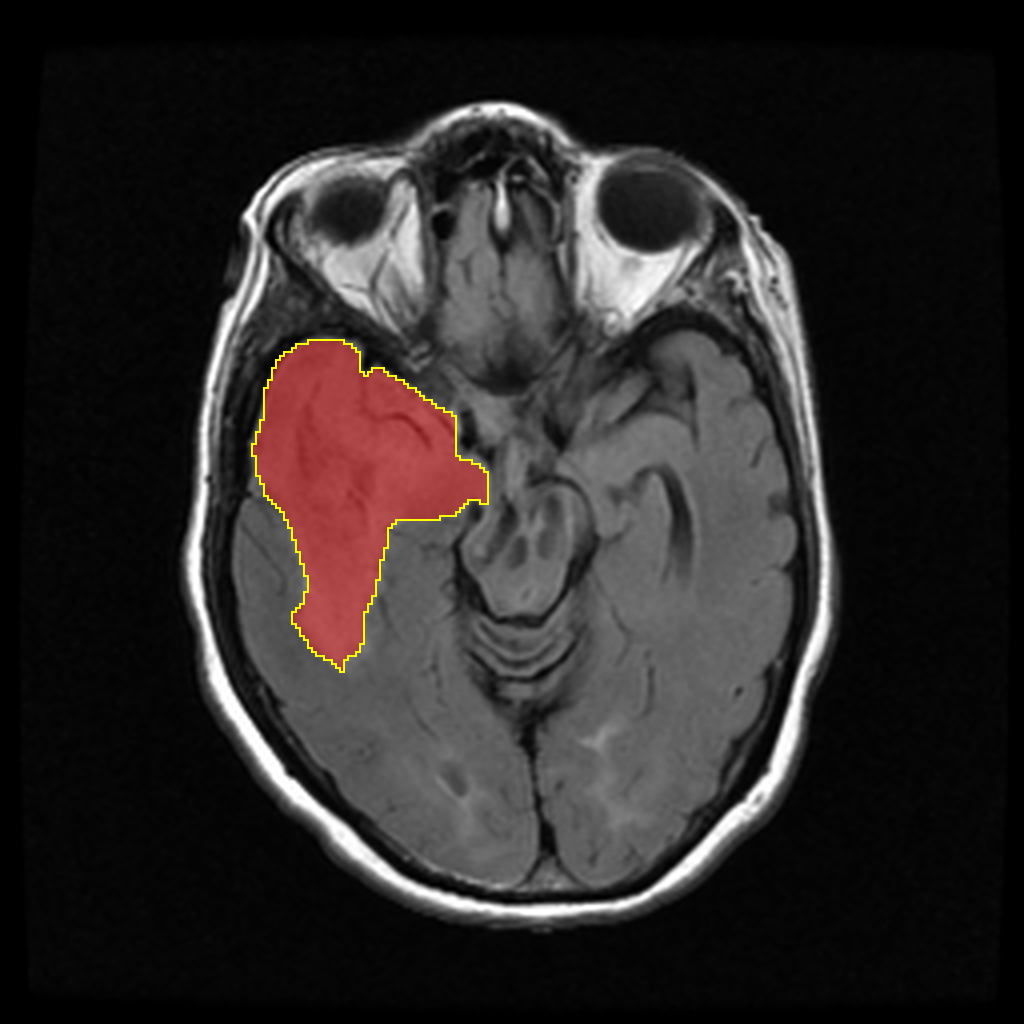}\\
{\small (a) Teacher view\\(overlay target)\par}
\end{minipage}\hfill
\begin{minipage}[t]{0.48\linewidth}
\centering
\includegraphics[width=\linewidth]{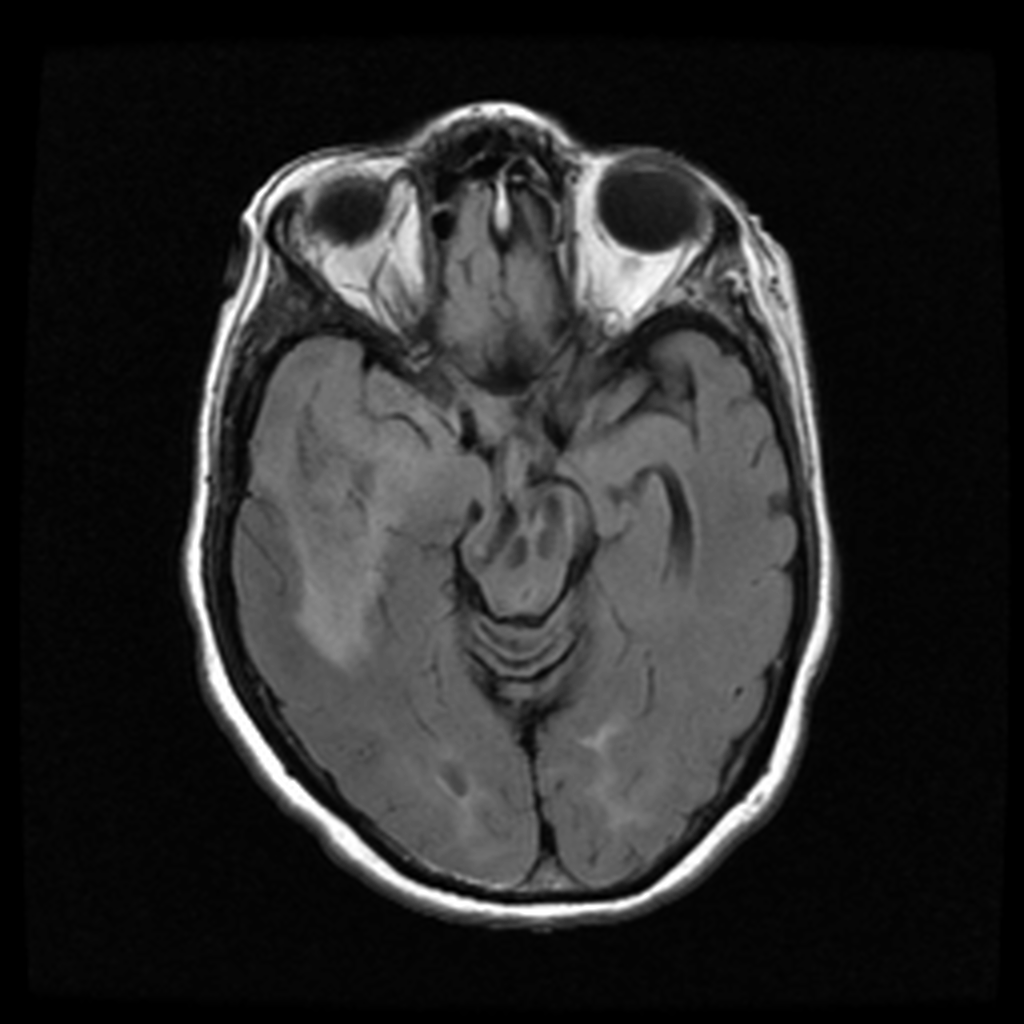}\\
{\small (b) Student view\\(raw image)\par}
\end{minipage}

\caption{
\textbf{Comparison between the teacher view and the student view.} 
The teacher view contains a semi-transparent target overlay for localization guidance, while the student view contains only the raw unannotated image.
}
\label{fig:teacher_student_overlay}
\end{figure}

This design creates two distinct visual views for the same sample. 
The \emph{teacher view} contains privileged localization information through the semi-transparent target overlay, whereas the \emph{student view} contains only the raw image without any annotation. 
During distillation, the teacher is explicitly instructed to use the overlay only for internal localization and to generate the final answer and reasoning chain \emph{as if it only sees the raw, unannotated image}. 
Figure~\ref{fig:teacher_student_overlay} illustrates the difference between the teacher view and the student view.
As a result, the distilled reasoning can remain grounded in the true target location while avoiding direct leakage of mask-, overlay-, or annotation-related cues.

\paragraph{System prompt design.}
The teacher prompt is designed to generate reasoning chains that are correct, image-grounded, and free from teacher-view leakage. 
In each API call, the teacher model receives a teacher-only overlay image, the hidden target identity, the target type, and the student-visible question. 
The textual prompt is organized into the following components.

\begin{enumerate}
    \item \textbf{Role definition.} 
    The system prompt defines the model as a teacher for medical visual reasoning data generation. 
    The teacher is instructed to generate concise supervision for latent reasoning training, including a final answer and a reasoning chain.

    \item \textbf{Teacher-only information usage.} 
    The teacher is allowed to use teacher-only information internally, including the overlay image and the hidden target identity, to localize the target region correctly. 
    However, the output must be written as if it were produced by a student model that can only observe the original unannotated image.

    \item \textbf{Annotation-leakage prevention.} 
    The prompt explicitly forbids any mention or implication of hidden guidance, special markings, annotations, masks, overlays, highlighted areas, segmentation maps, ROI, labels, ground truth, teacher-only metadata, or extra visual cues. 
    It also forbids statements suggesting that the correct target identity was provided beforehand.

    \item \textbf{Image-grounded medical reasoning.} 
    The teacher is required to base the answer and reasoning chain on visible image evidence, such as location, shape, boundary, extent, density or signal intensity, texture, and relationships to nearby structures. 
    For lesion or finding targets, the teacher may name the given target itself, but should not introduce unsupported diagnosis, malignancy or benignity, severity, prognosis, treatment, or other clinical claims unless they are directly supported by visible evidence.

    \item \textbf{Spatial convention.} 
    For spatial descriptions, the teacher must use image-space wording only, such as left, right, center, upper, and lower parts of the image. 
    Patient-space or radiological-convention wording, such as patient-left, patient-right, anatomical-left, or anatomical-right, is explicitly disallowed.

    \item \textbf{Output format.} 
    The teacher must return strict JSON with exactly two fields: \texttt{final\_answer} and \texttt{reasoning\_chain}. 
    Markdown, code fences, and additional explanations are not allowed.
\end{enumerate}

The sample-specific user prompt further injects the target identity, target type, question type, and student-visible question. 
It also appends a question-type-specific guidance block that specifies the required reasoning-step range and provides an in-context JSON example. 
This design separates persistent global constraints from per-sample control signals: the system prompt enforces annotation-free medical reasoning behavior, while the user prompt controls the current target, question, and expected reasoning granularity. 
If a generated response fails format checking, step-count validation, or leakage filtering, a retry prompt is appended to require regeneration from scratch under stricter reminders.

\paragraph{Question-type-aware prompting.}
To control the granularity of distilled reasoning chains, we use question-type-aware prompting. 
Each sample is assigned a question type either from the pre-generated metadata or, if unavailable, by rule-based classification from the question text. 
The BiomedParse question-generation script uses target-type- and modality-aware templates to generate five teacher-compatible question categories: \texttt{yesno}, \texttt{identify}, \texttt{location}, \texttt{describe}, and \texttt{reasoning}. 
These templates are adapted to heterogeneous biomedical targets, including organs, anatomical sub-structures, vessels, tissues, lesions or findings, retinal structures, and microscopy-level structures. 
During teacher generation, an additional \texttt{location\_choice} category is recognized for questions that ask the model to choose among coarse image regions such as left, right, or center.

For each question type, the prompt specifies an explicit reasoning-step target.

\begin{enumerate}
    \item \textbf{\texttt{yesno}.} 
    Simple visual confirmation questions use exactly one reasoning step. 
    The reasoning chain only needs to provide direct visual evidence supporting the yes/no answer.

    \item \textbf{\texttt{identify}.} 
    Identification questions also use exactly one reasoning step. 
    The teacher is expected to justify the recognized target based on its visible appearance and anatomical context.

    \item \textbf{\texttt{location\_choice}.} 
    Coarse spatial-choice questions use exactly one reasoning step. 
    The final answer must be a complete sentence rather than a bare word such as ``left'', ``right'', or ``center''.

    \item \textbf{\texttt{location}.} 
    Spatial localization questions use exactly two reasoning steps. 
    This allows the teacher to first identify the target region and then describe its image-space position relative to surrounding anatomy.

    \item \textbf{\texttt{describe}.} 
    Descriptive questions use two to three reasoning steps. 
    The reasoning chain should cover the target's appearance, shape, boundary, extent, and other visible characteristics.

    \item \textbf{\texttt{reasoning}.} 
    More complex visual interpretation questions use three to four reasoning steps. 
    These questions require a fuller analysis of the target's location, morphology, visual contrast, and relationship to nearby structures.
\end{enumerate}

\begin{figure*}[t]
\centering
\includegraphics[width=0.95\textwidth]{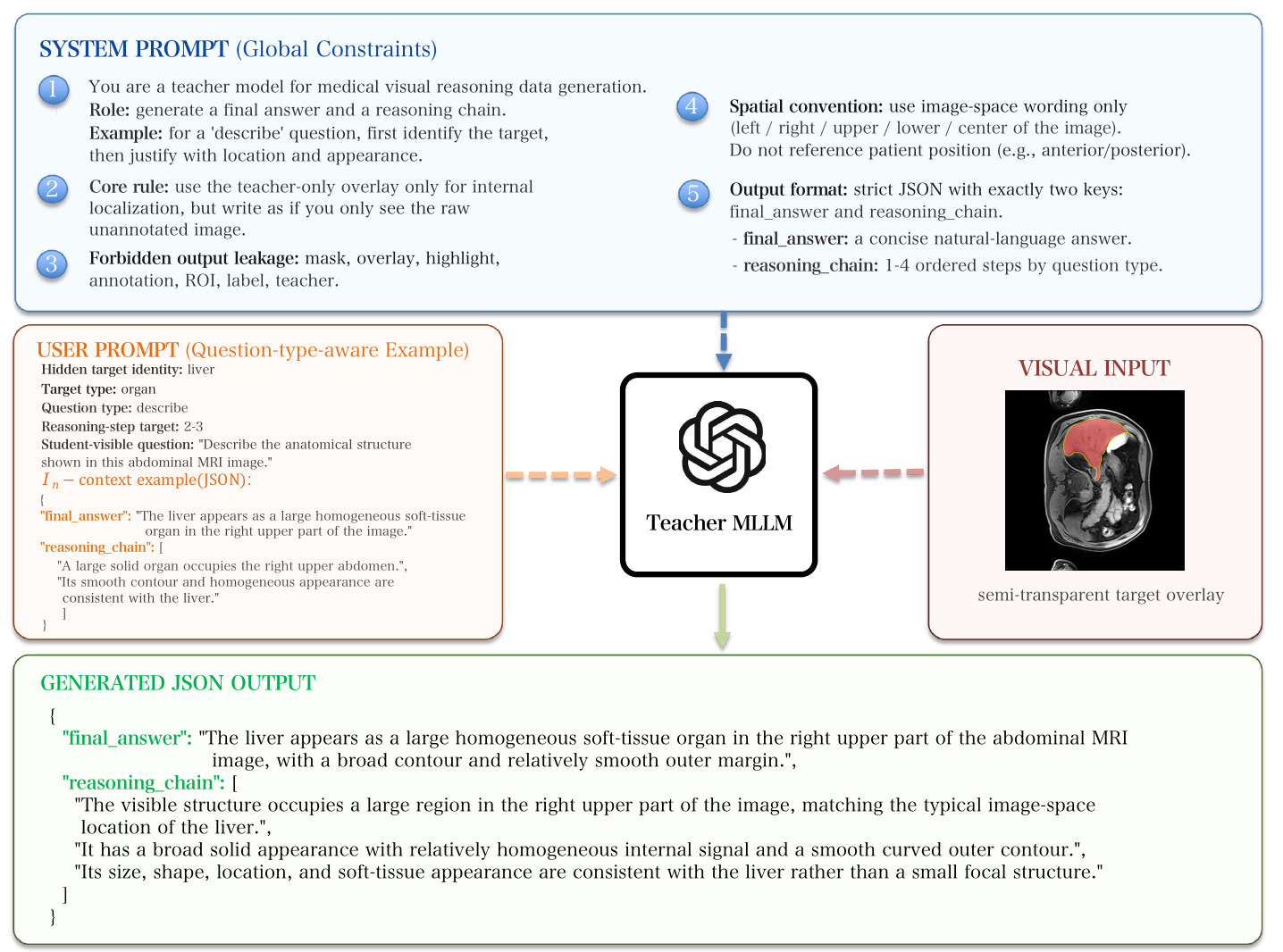}
\caption{
\textbf{Prompt design example for teacher distillation.} 
The teacher receives three sources of information in a single API call: a global system prompt that specifies the medical reasoning role, annotation-leakage constraints, image-space spatial convention, and strict JSON output format; a question-type-aware user prompt that provides the hidden target identity, target type, reasoning-step target, student-visible question, and an in-context JSON example; and a teacher-only overlay image that is used only for internal localization. 
The teacher MLLM then generates a student-view JSON output containing \texttt{final\_answer} and \texttt{reasoning\_chain}, written as if only the raw unannotated image were visible.
}
\label{fig:prompt_design_teacher_distillation}
\end{figure*}

Each question-type guidance block contains three elements: the normalized question type, the required reasoning-step range, and an in-context JSON example showing the desired output style. 
This makes both the output format and the reasoning depth explicit for the teacher model. 
After generation, the number of produced reasoning steps is checked against the allowed range for the corresponding question type. 
The validated step count is stored as \texttt{latent\_steps}, and the reasoning chain is stored as \texttt{text\_reasoning\_chain}. 
In this way, question-type-aware prompting directly determines the length of the latent reasoning supervision while preserving a consistent JSON format across all samples.
Figure~\ref{fig:prompt_design_teacher_distillation} illustrates a concrete prompt example used in our teacher-distillation pipeline.

\paragraph{Dataset-specific adaptations.}
For organs (normal anatomy), we additionally filter outputs against a pathology-style term list (e.g., \emph{lesion}, \emph{mass}, \emph{tumor}, \emph{calcification}, \emph{inflammation}) to prevent the teacher from hallucinating clinical findings on healthy organ data.
For other targets (which includes pathological objectives), pathology terms are permitted and expected.
Both variants share the same annotation-leakage filter and spatial-consistency checks.

\subsection{Multi-Round Quality Control}
\label{app:quality_control}

Each teacher-generated sample undergoes a six-round automated verification pipeline.
Samples failing any round are automatically retried with progressively stricter prompt constraints (up to 4 retries).

\begin{enumerate}
\item \textbf{JSON parse validation.}
Verify that the teacher output is valid JSON conforming to the required schema (\texttt{final\_answer}: string, \texttt{reasoning\_chain}: list of strings).

\item \textbf{Forbidden-term filtering.}
Scan both \texttt{final\_answer} and all entries of \texttt{reasoning\_chain} for any of the 30 annotation-leakage terms.
Any hit triggers rejection and retry.

\item \textbf{Pathology-style filtering (Organ only).}
For the normal-anatomy subset, check for inappropriate use of pathology terms (e.g., describing a healthy liver as having a ``lesion'' or ``mass'').
This filter is disabled for BiomedParse where pathological descriptions are valid.

\item \textbf{Location-mixed filtering.}
Detect mixing of patient-space and image-space spatial references (e.g., ``patient's right'' or ``anatomical left''), which violates our image-space-only convention.

\item \textbf{Step-count validation.}
Verify that $|\texttt{reasoning\_chain}|$ falls within the target range $[\texttt{min\_steps}, \texttt{max\_steps}]$ defined by the question type.

\item \textbf{Answer normalization.}
Standardize \texttt{final\_answer} format according to question type: identification answers shorter than 4 words are expanded into complete sentences (e.g., ``pancreas'' $\to$ ``The main organ shown is the pancreas.''); location answers are reformatted with explicit image-space phrasing; all answers are ensured to end with a period.
\end{enumerate}

\paragraph{Retry mechanism.}
Upon rejection, the sample is re-submitted with an appended ``strict rules'' prompt block that explicitly lists the violation and demands regeneration from scratch.
After 4 failed attempts, the sample is logged and excluded from the final dataset.

\paragraph{Human audit.}
After the automated pipeline completes, we conduct a 5\% random audit of the final dataset.
The observed annotation leakage rate is $<0.3\%$, confirming the effectiveness of our multi-round filtering.

\subsection{Dataset Statistics}
\label{app:data_stats}

\begin{figure*}[h]
\centering
\begin{minipage}{0.42\textwidth}
    \centering
    \includegraphics[width=\linewidth]{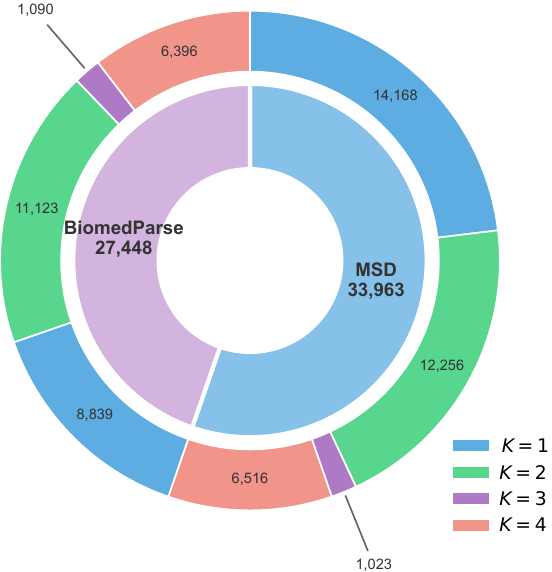}
\end{minipage}
\hfill
\begin{minipage}{0.55\textwidth}
    \centering
    \includegraphics[width=\linewidth]{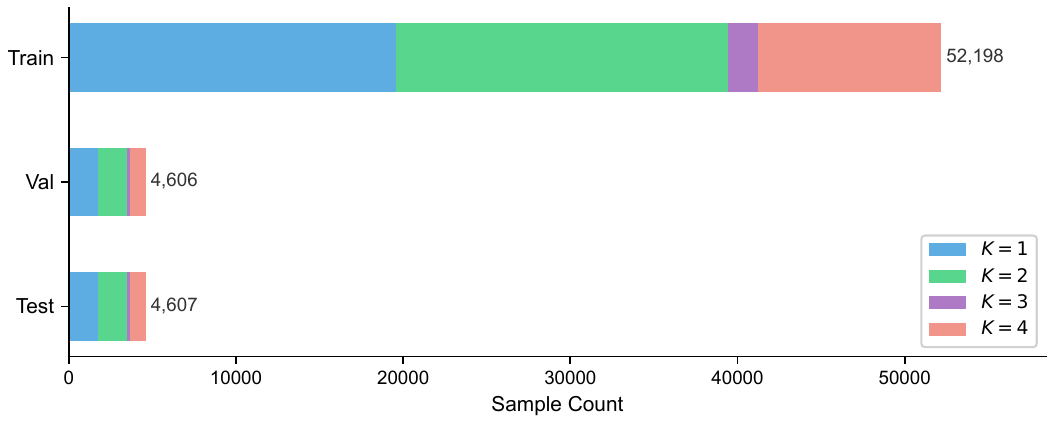}
\end{minipage}
\caption{\textbf{Dataset statistics.} \textbf{Left:} Sample distribution by source and reasoning depth $K$. \textbf{Right:} Train / validation / test split by reasoning depth $K$.}
\label{fig:data_stats}
\end{figure*}

The final VITAL dataset comprises \textbf{61{,}411} samples drawn from two public sources spanning 9 imaging modalities: CT, MRI, X-ray, ultrasound, endoscope, fundus photography, dermoscopy, OCT, and pathology.
Figure~\ref{fig:data_stats} and Tables~\ref{tab:data_subset}--\ref{tab:data_modal} provide a comprehensive breakdown by source, sub-set, train/val/test partition, and imaging modality.

\begin{table}[h]
\centering
\caption{\textbf{Per-sub-set sample counts by reasoning depth $K$.}}
\label{tab:data_subset}
\small
\resizebox{\columnwidth}{!}{%
\begin{tabular}{@{}lrrrrr@{}}
\toprule
\textbf{Sub-set} & \textbf{$K{=}1$} & \textbf{$K{=}2$} & \textbf{$K{=}3$} & \textbf{$K{=}4$} & \textbf{Total} \\
\midrule
MSD & 14{,}168 & 12{,}256 & 1{,}023 & 6{,}516 & 33{,}963 \\
\midrule
Radiography & 1{,}320 & 1{,}698 & 188 & 950 & 4{,}156 \\
amos22 & 1{,}124 & 1{,}439 & 132 & 730 & 3{,}425 \\
CAMUS & 647 & 834 & 27 & 488 & 1{,}996 \\
COVID-QU-Ex & 590 & 776 & 54 & 434 & 1{,}854 \\
ACDC & 580 & 747 & 26 & 393 & 1{,}746 \\
MMs & 580 & 746 & 29 & 388 & 1{,}743 \\
CXR\_Masks\_and\_Labels & 497 & 599 & 52 & 361 & 1{,}509 \\
NeoPolyp & 444 & 609 & 52 & 368 & 1{,}473 \\
kits23 & 469 & 551 & 69 & 338 & 1{,}427 \\
ISIC & 358 & 339 & 117 & 261 & 1{,}075 \\
PanNuke & 352 & 445 & 12 & 263 & 1{,}072 \\
PolypGen & 316 & 390 & 32 & 220 & 958 \\
LGG & 293 & 278 & 107 & 232 & 910 \\
FH-PS-AOP & 274 & 368 & 21 & 215 & 878 \\
LIDC-IDRI & 230 & 342 & 42 & 177 & 791 \\
BreastUS & 187 & 236 & 42 & 131 & 596 \\
QaTa-COV19 & 97 & 133 & 11 & 73 & 314 \\
G1020 & 73 & 112 & 12 & 68 & 265 \\
siim-acr-pneumothorax & 91 & 81 & 22 & 58 & 252 \\
UWaterlooSkinCancer & 69 & 81 & 24 & 53 & 227 \\
COVID-19\_CT & 61 & 101 & 7 & 55 & 224 \\
OCT-CME & 74 & 83 & 2 & 46 & 205 \\
REFUGE & 61 & 79 & 5 & 55 & 200 \\
GlaS & 45 & 44 & 5 & 29 & 123 \\
LiverUS & 7 & 12 & 0 & 10 & 29 \\
\midrule
\textbf{Total} & \textbf{23{,}007} & \textbf{23{,}379} & \textbf{2{,}113} & \textbf{12{,}912} & \textbf{61{,}411} \\
\bottomrule
\end{tabular}%
}
\end{table}

\begin{table}[h]
\centering
\caption{\textbf{Sample distribution by imaging modality and reasoning depth $K$.}}
\label{tab:data_modal}
\small
\setlength{\tabcolsep}{4pt}
\begin{adjustbox}{max width=\linewidth}
\begin{tabular}{@{}lrrrrr@{}}
\toprule
\textbf{Modality} & \textbf{$K{=}1$} & \textbf{$K{=}2$} & \textbf{$K{=}3$} & \textbf{$K{=}4$} & \textbf{Total} \\
\midrule
CT & 13{,}622 & 12{,}470 & 1{,}071 & 6{,}618 & 33{,}781 \\
MRI & 3{,}883 & 3{,}990 & 364 & 2{,}211 & 10{,}448 \\
X-Ray & 2{,}595 & 3{,}287 & 327 & 1{,}876 & 8{,}085 \\
Ultrasound & 1{,}115 & 1{,}450 & 90 & 844 & 3{,}499 \\
Endoscope & 760 & 999 & 84 & 588 & 2{,}431 \\
Dermoscopy & 427 & 420 & 141 & 314 & 1{,}302 \\
Pathology & 397 & 489 & 17 & 292 & 1{,}195 \\
Fundus & 134 & 191 & 17 & 123 & 465 \\
OCT & 74 & 83 & 2 & 46 & 205 \\
\midrule
\textbf{Total} & \textbf{23{,}007} & \textbf{23{,}379} & \textbf{2{,}113} & \textbf{12{,}912} & \textbf{61{,}411} \\
\bottomrule
\end{tabular}
\end{adjustbox}
\end{table}

\paragraph{Source diversity.}
The MSD portion (33{,}963 samples) is derived from the Medical Segmentation Decathlon, covering 10 organ classes from 3D CT/MRI volumes with dense axial-slice sampling.
The BiomedParse portion (27{,}448 samples, 25 sub-sets) extends coverage to pathological findings and diverse biomedical structures across 9 imaging modalities (Table~\ref{tab:data_modal}).
CT and MRI together account for 72.0\% of all samples, providing a strong foundation in cross-sectional imaging; X-ray (13.2\%) and ultrasound (5.7\%) offer complementary image projection and real-time modalities; and the remaining modalities (endoscope, dermoscopy, pathology, fundus, and OCT) collectively contribute 9.1\%, ensuring the model is exposed to the full breadth of clinical imaging.

\paragraph{Reasoning depth distribution.}
Single-step samples ($K{=}1$, 37.5\%) and two-step samples ($K{=}2$, 38.1\%) together account for the majority, reflecting the prevalence of identification and localization questions.
Multi-step reasoning ($K{=}3{-}4$, 24.5\%) is concentrated in descriptive and analytical questions requiring richer clinical rationale.
This distribution naturally aligns with our curriculum learning strategy (\S\ref{sec:training}): Phase~1 and~2 can leverage the abundant $K{\leqslant}2$ samples, while Phase~3 trains on the full spectrum.

\paragraph{Comparison with prior datasets.}
To our knowledge, this is the largest dataset specifically designed for training latent/chain-of-thought reasoning in the medical vision domain.
Existing medical reasoning datasets are significantly smaller: MedThink~\citep{gai2025medthink} provides rationale annotations for $\sim$9.5K QA pairs (R-RAD: 3{,}515 + R-SLAKE: 5{,}980), and M3CoTBench~\citep{jiang2026m3cotbench} offers only 1{,}079 samples as an evaluation benchmark.
Our dataset exceeds these efforts by $6{\times}$--$57{\times}$ in scale, while additionally providing aligned ROI visual features for every sample, a unique annotation absent from all prior medical reasoning datasets.

\subsection{Adaptive ROI Feature Extraction}
\label{app:roi_extraction}

For each training sample, we extract a $d_v$-dimensional ROI feature from a frozen medical vision encoder as the supervision target for $\mathcal{L}_{\text{visual}}$.
This section describes the encoder selection, the adaptive extraction strategy, and the hyperparameter tuning that together ensure high-quality visual supervision signals.

\subsubsection{Vision Encoder Selection}
\label{app:encoder_selection}

We evaluate two biomedical vision encoders as candidates for ROI feature extraction:
\textbf{BiomedCLIP}~\citep{zhang2023biomedclip}, a CLIP-based model pretrained on 15M PubMed image--text pairs, and
\textbf{MedSigLIP}~\citep{sellergren2025medgemma}, a SigLIP variant medically tuned by Google as part of the MedGemma family.
Table~\ref{tab:encoder_compare} summarizes the key architectural differences.

\begin{table}[h]
\centering
\caption{\textbf{Comparison of candidate medical vision encoders.}}
\label{tab:encoder_compare}
\small
\resizebox{\columnwidth}{!}{%
\begin{tabular}{@{}lcc@{}}
\toprule
\textbf{Property} & \textbf{BiomedCLIP} & \textbf{MedSigLIP} \\
\midrule
Input resolution & $224{\times}224$ & $448{\times}448$ \\
Patch size & 16 & 14 \\
Patch grid & $14{\times}14 = 196$ & $32{\times}32 = 1{,}024$ \\
Hidden dim ($d_v$) & 768 & 1{,}152 \\
CLS token & Yes & No (all spatial) \\
\bottomrule
\end{tabular}%
}
\end{table}

The critical advantage of MedSigLIP is its $5.2{\times}$ higher spatial resolution: 1{,}024 patches versus 196.
For a typical medical image (e.g., $1{,}024{\times}1{,}024$ pixels), each BiomedCLIP patch covers ${\sim}73{\times}73$ pixels, whereas each MedSigLIP patch covers only ${\sim}14{\times}14$ pixels.
This dramatically improves the fidelity of mask-aligned patch selection: a small lesion occupying 2\% of the image area would align with ${\sim}4$ BiomedCLIP patches but ${\sim}20$ MedSigLIP patches, yielding a far richer and more discriminative ROI feature after mean pooling.
Additionally, SigLIP's architecture produces only spatial patch tokens (no CLS token), which simplifies the mask-to-patch alignment pipeline.
We therefore adopt \textbf{MedSigLIP} (SigLIP-SO400M/14-448) as our frozen vision encoder throughout all experiments.
This choice is further validated by end-to-end ablation (\S\ref{app:ablation_encoder}), where MedSigLIP outperforms both BiomedCLIP and the backbone's own encoder by a large margin on downstream accuracy.

\begin{figure*}[t]
\centering
\includegraphics[width=\textwidth]{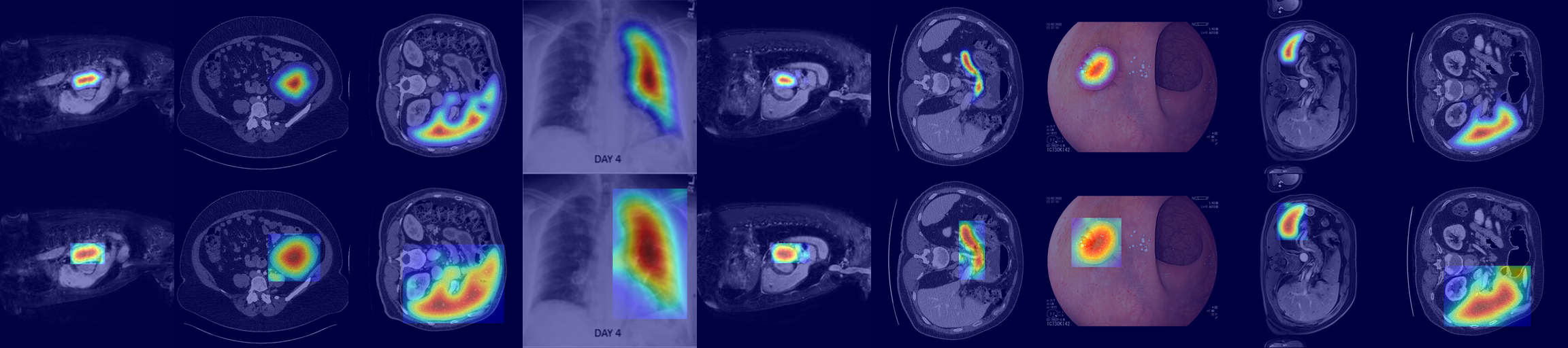}
\caption{\textbf{Qualitative comparison of patch-level activation heatmaps between MedSigLIP and BiomedCLIP.} MedSigLIP (\textbf{top row}) produces sharply localized activations concentrated on the target region, while BiomedCLIP (\textbf{bottom row}) exhibits diffuse, spatially imprecise responses due to its coarser $14{\times}14$ patch grid.}
\label{fig:encoder_compare}
\end{figure*}

Figure~\ref{fig:encoder_compare} provides a qualitative comparison on representative samples spanning CT, X-ray, endoscopy, and dermoscopy modalities.
For each sample, we visualize the patch-level activation heatmap (cosine similarity between each patch feature and the mean ROI feature, overlaid on the original image).
MedSigLIP's activations are sharply concentrated on the target anatomical structure with minimal background noise, whereas BiomedCLIP's lower-resolution grid produces diffuse, spatially imprecise activations that often spread into surrounding tissue.
This contrast is especially pronounced for small or thin structures (e.g., vessels, polyps), where BiomedCLIP's 196-patch grid lacks the granularity to isolate the target from its context.

\subsubsection{Extraction Pipeline}
\label{app:extraction_pipeline}

We extract patch-level features by hooking the output of the last vision encoder block, yielding a $1{,}024 \times 1{,}152$ feature map per image.
The extraction follows an \emph{adaptive strategy} governed by two hyperparameters: the \textbf{area threshold} $T$ and the \textbf{crop margin ratio} $P$.

\paragraph{Full-image pathway (mask ratio $\geqslant T$).}
When the target structure is large relative to the image, we feed the entire image to MedSigLIP.
The segmentation mask is downsampled to the $32{\times}32$ patch grid via area interpolation (binarized at 0.3), and the ROI feature is computed as the mean of all mask-aligned patch features.

\paragraph{Tight-crop pathway (mask ratio $< T$).}
When the target is small, a full-image approach would dilute the ROI signal among overwhelming background patches.
Instead, we tightly crop the ROI using the mask bounding box expanded by a margin ratio $P$ on each side, pad to a square with zero-fill, and re-extract patch features from the cropped image.
A validity mask distinguishes real content from padding regions, and only valid patches contribute to the mean-pooled ROI feature.

\paragraph{Post-processing.}
All ROI features are $\ell_2$-normalized to the unit hypersphere.
For each sample, we additionally store the top-5 patches (by cosine similarity with the ROI feature) along with their grid positions, enabling downstream visualization of which spatial locations contribute most to the feature representation.
Each sample's output is saved as a \texttt{.pt} file.

\subsubsection{Hyperparameter Grid Search}
\label{app:roi_grid_search}

The two hyperparameters, area threshold $T$ and crop margin ratio $P$, jointly determine how and when the extraction strategy switches.
We conduct a grid search over $T \in \{0.05, 0.10, 0.20\}$ and $P \in \{0.0, 0.03, 0.05, 0.10\}$ on a random subset of 1,000 samples, evaluating three complementary metrics:

\begin{itemize}
\item \textbf{Coverage (\%)}. Fraction of the $32{\times}32$ patch grid covered by the ROI mask after downsampling.
Higher values indicate denser spatial supervision.
\item \textbf{Intensity}. Mean $\ell_2$-norm of patch activations within the ROI region, reflecting feature activation strength.
\item \textbf{SNR}. Ratio of mean ROI activation to mean background activation, measuring how well the ROI stands out from the background.
\end{itemize}

\begin{figure*}[t]
\centering
\includegraphics[width=\textwidth]{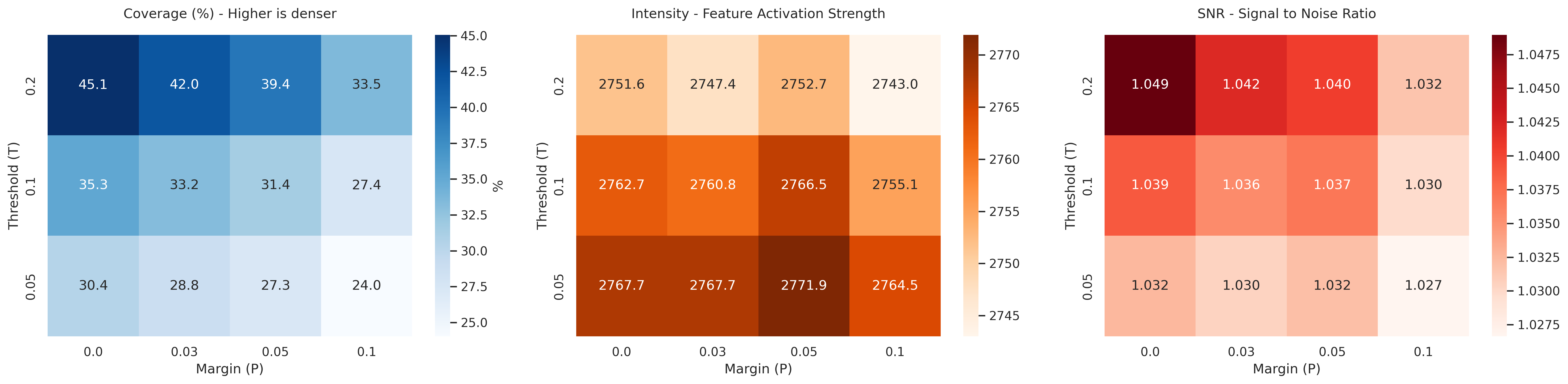}
\caption{\textbf{Grid search over area threshold $T$ and crop margin ratio $P$ for adaptive ROI feature extraction.}
\textbf{Left}: Coverage, where higher $T$ triggers full-image extraction for more samples, increasing patch coverage.
\textbf{Center}: Feature activation intensity within the ROI.
\textbf{Right}: Signal-to-noise ratio (ROI vs.\ background activation).
The selected configuration $T{=}0.20$, $P{=}0.05$ (highlighted) achieves the best balance: moderate coverage (39.4\%), near-peak intensity (2{,}752.7), and the highest SNR (1.040) among configurations with $T{\geqslant}0.10$.}
\label{fig:roi_grid_search}
\end{figure*}

Figure~\ref{fig:roi_grid_search} presents the results as three heatmaps.
Several trends emerge:
\begin{itemize}
\item \textbf{Coverage increases monotonically with $T$}. A higher threshold routes more samples to the full-image pathway, which naturally produces denser patch grids.
However, excessively high coverage (e.g., $T{=}0.20$, $P{=}0.0$: 45.1\%) comes at the cost of including many background patches.
\item \textbf{Intensity peaks at moderate $T$}. The tight-crop pathway ($T{=}0.05$) yields the highest raw activation (2{,}767.7--2{,}771.9) because cropped images concentrate the target, but this advantage diminishes as background patches are included.
\item \textbf{SNR favors higher $T$ with small $P$}. The SNR peaks around $T{=}0.20$ (1.032--1.049), indicating that the full-image pathway preserves contextual contrast between ROI and background better than aggressive cropping.
\end{itemize}

We select $\mathbf{T{=}0.20}$ and $\mathbf{P{=}0.05}$ as the final configuration, as it achieves the best trade-off:
competitive intensity (2{,}752.7), the highest SNR among non-trivial coverage levels (1.040), and sufficient coverage (39.4\%) for robust mean pooling.
This configuration is used for all 61{,}411 samples in the final dataset.

\section{Implementation Details}
\label{app:impl_details}

This appendix provides the complete training configuration, per-module parameter breakdown, and evaluation protocol details for reproducibility.

\subsection{Model Architecture}
\label{app:model_arch}

VITAL is built on top of \textbf{Qwen3-VL-8B-Instruct}~\citep{bai2025qwen3vltechnicalreport}, a multimodal large language model comprising a vision encoder and a language backbone.
After accounting for weight tying between \texttt{embed\_tokens} and \texttt{lm\_head}, the de-duplicated parameter count is 8.14B.
Table~\ref{tab:backbone_config} summarizes the architectural configuration.

\begin{table}[h]
\centering
\caption{\textbf{Backbone architecture configuration.}}
\label{tab:backbone_config}
\small
\begin{tabular}{@{}lcc@{}}
\toprule
\textbf{Property} & \textbf{LLM} & \textbf{Vision Encoder} \\
\midrule
Hidden size & 4{,}096 & 1{,}152 \\
Layers & 36 & 27 \\
Attention heads & 32 & 16 \\
KV heads (GQA) & 8 & --- \\
Intermediate size & 12{,}288 & --- \\
Patch size & --- & 16 \\
Vocab size & \multicolumn{2}{c}{151{,}936} \\
\midrule
Parameters & 7{,}568M & 576M \\
\bottomrule
\end{tabular}
\end{table}

\subsection{Trainable Modules}
\label{app:trainable_modules}

Table~\ref{tab:params} gives the full parameter breakdown.
Both the vision encoder and the LLM backbone are \emph{frozen}; trainability is introduced through LoRA adapters and three auxiliary modules that serve as training-time scaffolding.

\begin{table}[h]
\centering
\caption{\textbf{Parameter statistics of VITAL.} \emph{Scaffolding} modules (marked with $\dagger$) are discarded at inference. \colorbox{orange!15}{Highlighted} rows indicate trainable modules.}
\label{tab:params}
\small
\resizebox{\columnwidth}{!}{%
\begin{tabular}{@{}lrrr@{}}
\toprule
\textbf{Module} & \textbf{\#Params} & \textbf{\#Trainable} & \textbf{Status} \\
\midrule
Vision Encoder & 576.4M & 0 & Frozen \\
LLM Backbone & 7{,}568.4M & 0 & Frozen \\
\rowcolor{orange!15} \quad + LoRA ($r{=}64$, $\alpha{=}128$) & 174.6M & 174.6M & Trained \\
\rowcolor{orange!15} Visual Projector$^\dagger$ & 21.5M & 21.5M & Trained \\
\rowcolor{orange!15} Projection $\pi_{\text{in}}$$^\dagger$ & 8.4M & 8.4M & Trained \\
\rowcolor{orange!15} Text Decoder (Qwen3-1.7B)$^\dagger$ & 1{,}720.6M & 1{,}720.6M & Trained \\
\midrule
\textbf{Total (de-duplicated)} & \textbf{10{,}069.9M} & \textbf{1{,}925.1M (19.1\%)} & \\
\bottomrule
\end{tabular}%
}
\end{table}

\paragraph{Visual Projector.}
The visual projector maps the LLM hidden state ($d{=}4{,}096$) to the MedSigLIP feature space ($d_v{=}1{,}152$) using a norm-stabilized residual MLP (21.51M parameters):
\[
\texttt{VP}(z) = \text{LN}_{d_v}\!\Big(W_2\big(\text{GELU}(W_1\,\text{LN}_{d}(z)) + z\big)\Big),
\]
where $W_1 \in \mathbb{R}^{4096 \times 4096}$ (16.78M), $W_2 \in \mathbb{R}^{1152 \times 4096}$ (4.72M), and a dropout of 0.1 is applied after GELU.
The input LayerNorm eliminates the norm imbalance between LLM hidden states and vision features; the residual connection preserves representational richness; and the output LayerNorm re-scales to the $\ell_2$-normalized feature space.

\paragraph{Projection layer $\pi_{\text{in}}$.}
A single linear layer $\pi_{\text{in}} \in \mathbb{R}^{2048 \times 4096}$ with bias projects the LLM hidden state to the text decoder's input dimension (8.39M parameters).

\paragraph{Auxiliary Text Decoder.}
We use Qwen3-1.7B as the auxiliary text decoder, fully fine-tuned during training.
It is an independent causal language model with its own embedding layer and language model head (weight-tied), sharing no parameters with the backbone.
Table~\ref{tab:decoder_config} summarizes its configuration.

\begin{table}[h]
\centering
\caption{\textbf{Auxiliary text decoder (Qwen3-1.7B) configuration.}}
\label{tab:decoder_config}
\small
\begin{tabular}{@{}lr@{}}
\toprule
\textbf{Property} & \textbf{Value} \\
\midrule
Hidden size & 2{,}048 \\
Layers & 28 \\
Attention heads & 16 \\
KV heads (GQA) & 8 \\
Intermediate size & 6{,}144 \\
Vocab size & 151{,}936 \\
\texttt{embed\_tokens} & 311.16M \\
Transformer layers & 1{,}409.41M \\
\texttt{lm\_head} & weight-tied \\
\midrule
\textbf{Total} & \textbf{1{,}720.57M} \\
\bottomrule
\end{tabular}
\end{table}

\paragraph{External vision encoder for ROI features.}
ROI visual features are extracted by a frozen MedSigLIP encoder (SigLIP-SO400M/14-448)~\citep{sellergren2025medgemma}, which accepts $448{\times}448$ images and produces a $32{\times}32{=}1{,}024$ patch grid with hidden dimension $d_v{=}1{,}152$.
This encoder is used \emph{only} for pre-extracting supervision targets and is not part of the VITAL model at either training or inference time.
The adaptive extraction strategy is detailed in \S\ref{app:roi_extraction}.

\subsection{Training Configuration}
\label{app:training_config}

Table~\ref{tab:train_config} lists the complete set of training hyperparameters.

\begin{table}[h]
\centering
\caption{\textbf{Training hyperparameters.}}
\label{tab:train_config}
\small
\begin{tabular}{@{}lr@{}}
\toprule
\textbf{Hyperparameter} & \textbf{Value} \\
\midrule
Optimizer & AdamW \\
$\beta_1$, $\beta_2$ & 0.9, 0.999 \\
Learning rate & $2 \times 10^{-5}$ \\
Warmup ratio & 0.05 \\
Weight decay & 0.01 \\
Max gradient norm & 1.0 \\
Precision & bf16 \\
$\lambda_{\text{text}}$ & 1.0 \\
$\lambda_{\text{visual}}$ & 0.1 \\
LoRA dropout & 0.05 \\
\midrule
\multicolumn{2}{@{}l}{\textbf{Curriculum schedule}} \\
\midrule
Phase 1 ($K{=}1$) & 5 epochs, $\sim$19.6K samples \\
Phase 2 ($K{\leqslant}2$) & 5 epochs, $\sim$39.4K samples \\
Phase 3 (all $K$) & 10 epochs, $\sim$52.2K samples \\
\midrule
Hardware & 6$\times$ NVIDIA RTX PRO 6000 \\
Total training time & $\sim$96 hours \\
\bottomrule
\end{tabular}
\end{table}

Each curriculum phase warm-starts from the previous phase's checkpoint.
The loss weights $\lambda_{\text{text}}{=}1.0$ and $\lambda_{\text{visual}}{=}0.1$ are kept constant across all phases; sensitivity analysis is provided in \S\ref{app:loss_weight}.

\subsection{Evaluation Protocol}
\label{app:eval_protocol}

\paragraph{In-domain testsets (MSD, BiomedParse).}
We use stratified held-out test splits (4{,}607 samples total; see Figure~\ref{fig:data_stats}) and report overall accuracy. In evaluation, all the question-answer pairs are turned into closed-ended questions.
The predicted answer is matched against the ground truth after lowercasing and whitespace normalization.

\paragraph{VQA-RAD, PathVQA.}
We follow the standard evaluation protocols established by each benchmark.
For VQA-RAD~\citep{lau2018dataset} and PathVQA~\citep{he2020pathvqa30000questionsmedical}, we report accuracy for closed-ended subsets and \textbf{token-level F1} for open-ended subsets.
Token-level F1 is computed by treating the predicted and ground-truth answers as bags of tokens (after lowercasing and whitespace tokenization), then calculating the harmonic mean of token-level precision and recall.

\paragraph{PMC-VQA.}
PMC-VQA~\citep{zhang2024pmcvqavisualinstructiontuning} is a multiple-choice benchmark; we report accuracy on the test split.

\paragraph{GEMeX-RMCoT.}
GEMeX-RMCoT~\citep{Liu_2025} evaluates medical visual grounding through reasoning-based multiple-choice questions that require attending to specific anatomical regions.
We report accuracy following the official evaluation script.

\paragraph{Held-out in-house testset.}
To mitigate evaluation bias caused by training-set contamination leading to inflated accuracy and token-F1 scores, we introduce a private held-out evaluation set.
This testset consists of 1{,}000 VQA pairs derived from brain neuroimaging data (MRI and PET scans), covering questions about anatomical identification, lesion localization, and clinical reasoning.
All questions are formulated as closed-ended (multiple-choice or yes/no); we report accuracy.
None of the evaluated baselines have been exposed to this data during training, ensuring a fair comparison.

\section{More Experiments}
\label{app:more_experiments}

\subsection{VITAL vs. Matched Verbalized-CoT Fine-Tuning}
\label{app:verbalized_cot}

To distinguish the contribution of VITAL's latent reasoning framework from that of the training data, we construct a matched verbalized-CoT baseline.
Specifically, we fine-tune Qwen3-VL-8B with LoRA using the same 61K training samples and the same LoRA configuration as VITAL, but require the model to generate each reasoning chain explicitly as text tokens before producing the final answer.
This matched baseline should not be confused with Qwen3-VL-8B-Thinking in the main comparison and efficiency analysis: the latter is a zero-shot explicit-CoT model that is not fine-tuned on our 61K dataset.
We additionally include a direct-answer LoRA baseline, the zero-shot Qwen3-VL-8B backbone, and the zero-shot Qwen3-VL-8B-Thinking model.
This comparison controls for both the backbone and training data, isolating the effect of representing the reasoning process in explicit text versus continuous latent states with visual-semantic dual supervision.

\begin{table}[h]
\centering
\small
\caption{\textbf{Comparison with matched verbalized-CoT fine-tuning.} The matched verbalized-CoT baseline and VITAL use the same Qwen3-VL-8B backbone, LoRA configuration, and 61K dataset. Qwen3-VL-8B-Thinking instead performs zero-shot explicit CoT without training on this dataset. In-D. denotes average accuracy on the two in-domain testsets, and RAD denotes closed-ended accuracy on VQA-RAD. Accuracy values are percentages.}
\label{tab:verbalized_cot}
\setlength{\tabcolsep}{3pt}
\resizebox{\columnwidth}{!}{%
\begin{tabular}{@{}lccccc@{}}
\toprule
\textbf{Method} & \textbf{Data} & \textbf{In-D.} & \textbf{RAD} & \textbf{In-H.} & \textbf{Latency} \\
\midrule
Qwen3-VL-8B (zero-shot) & --  & 33.97 & 67.63 & 52.60 & 286\,ms \\
$+$ LoRA (direct answer) & 61K & 45.47 & 58.31 & 56.60 & 325\,ms \\
$+$ LoRA (verbalized CoT, matched) & 61K & 67.87 & 62.53 & 54.10 & 20.6\,s \\
Qwen3-VL-8B-Thinking (zero-shot) & -- & 38.35 & 64.52 & 55.30 & 34.1\,s \\
\rowcolor{orange!15}
\textbf{VITAL} & \textbf{61K} & \textbf{81.08} & \textbf{74.28} & \textbf{80.50} & \textbf{353\,ms} \\
\bottomrule
\end{tabular}%
}
\end{table}

As shown in Table~\ref{tab:verbalized_cot}, VITAL substantially outperforms the matched verbalized-CoT baseline on all three representative evaluations, with absolute gains of 13.21 points on the in-domain testsets, 11.75 points on VQA-RAD, and 26.40 points on the held-out in-house testset.
Because the two methods share the same backbone, LoRA configuration, and training samples, these results show that the gains cannot be attributed to the 61K dataset alone; the latent reasoning mechanism and its dual supervision provide a clear advantage over verbalizing the same reasoning chains.

The verbalized-CoT baseline improves in-domain accuracy over direct-answer LoRA (67.87 vs.\ 45.47), but does not transfer as reliably beyond the training distribution.
On VQA-RAD, it scores 62.53, below the zero-shot backbone's 67.63; on the held-out in-house testset, it also underperforms direct-answer LoRA (54.10 vs.\ 56.60).
This pattern suggests that explicit rationale fine-tuning may overfit to the linguistic form of the training reasoning chains, whereas VITAL's visual-semantic constraints preserve stronger visual grounding and generalization.

VITAL is also substantially more efficient.
The verbalized-CoT model requires 20.6 seconds to generate its reasoning tokens before completing an answer, while VITAL performs four single-state recurrent reasoning steps and completes inference in 353\,ms, corresponding to a $58.4\times$ speedup.
The zero-shot Qwen3-VL-8B-Thinking model is even slower at 34.1 seconds and remains less accurate than the matched verbalized-CoT model on the in-domain and in-house evaluations.
Together, these results demonstrate that VITAL's advantage arises from the reasoning framework rather than data exposure alone: continuous reasoning with visual-semantic dual supervision improves both predictive performance and inference efficiency over both zero-shot Thinking CoT and matched verbalized-CoT fine-tuning.

\subsection{Ablation on Loss Weight Sensitivity}
\label{app:loss_weight}

\begin{figure}[h]
\centering
\includegraphics[width=0.75\columnwidth]{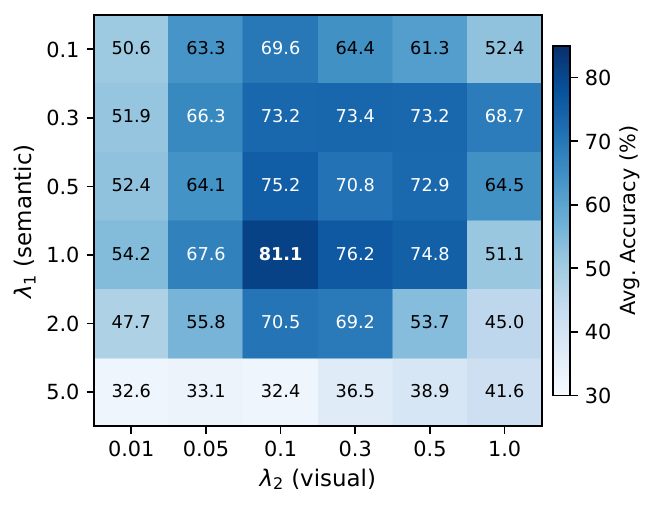}
\caption{\textbf{Loss weight sensitivity heatmap.} Each cell shows in-domain avg.\ accuracy (\%). The best setting ($\lambda_1{=}1.0$, $\lambda_2{=}0.1$) is used as default.}
\label{fig:loss_weight}
\end{figure}

As a complement to the dual supervision ablation in \S\ref{par:ablation_dual} (Table~\ref{tab:ablation_dual}), we perform a grid search over the loss weights $\lambda_1$ (semantic) and $\lambda_2$ (visual) to assess sensitivity.
Figure~\ref{fig:loss_weight} reports average accuracy across in-domain testsets for each $(\lambda_1, \lambda_2)$ combination.
The optimal setting is $\lambda_1{=}1.0$, $\lambda_2{=}0.1$ (81.08\%).
Performance is more sensitive to $\lambda_2$: moderate values (0.05--0.3) consistently outperform extremes, as too-large $\lambda_2$ forces latent states toward the visual subspace at the expense of reasoning capacity.
For $\lambda_1$, values in $[0.3, 1.0]$ are relatively stable, but excessive semantic weight ($\lambda_1{\geqslant}2.0$) degrades performance, likely due to the auxiliary decoder dominating gradient flow and hindering task-oriented learning.
We also observe that $\lambda_1{=}5.0$ collapses performance regardless of $\lambda_2$, confirming that balanced dual supervision is critical.

\subsection{Ablation on Visual Projector Architecture}
\label{app:vp_arch}

We compare five architectural choices for the visual projector (VP) that maps latent states to the medical vision encoder's feature space.
Results are reported in Table~\ref{tab:vp_arch}.
A plain linear layer (a) performs poorly (27.72\%), confirming that bridging the LLM hidden space to the vision feature space requires non-linear capacity.
Adding an MLP (b) doubles accuracy, and LayerNorm (c) provides a further +18.65 gain by resolving the norm imbalance between the two spaces, consistent with our motivation in \S\ref{sec:dual_supervision}.
The residual connection (d) yields the largest single-component boost (+12.25 over c), as it preserves the representational richness of latent states while the non-linear pathway learns a modality-bridging correction.
Surprisingly, the Transformer Block (e) collapses to 46.36\%, likely because the self-attention over a single token adds no useful inductive bias while introducing optimization instability.

\begin{table}[h]
\centering
\small
\caption{\textbf{Ablation on visual projector architecture.} In-domain avg.\ accuracy (\%) is reported.}
\label{tab:vp_arch}
\setlength{\tabcolsep}{4pt}
\begin{tabular}{@{}lccccc@{}}
\toprule
\textbf{Variant} & \textbf{MLP} & \textbf{LN} & \textbf{Res.} & \textbf{Attn.} & \textbf{In-D.} \\
\midrule
(a) Linear             &            &            &            &            & 27.72 \\
(b) MLP                & \checkmark &            &            &            & 50.18 \\
(c) MLP + LN           & \checkmark & \checkmark &            &            & 68.83 \\
\rowcolor{orange!15}
(d) \textbf{Res. MLP + LN (ours)} & \checkmark & \checkmark & \checkmark &            & \textbf{81.08} \\
(e) Transformer Block  & \checkmark & \checkmark & \checkmark & \checkmark & 46.36 \\
\bottomrule
\end{tabular}
\end{table}

\subsection{Ablation on Vision Encoder for Visual Supervision}
\label{app:ablation_encoder}

A central claim of VITAL is that visual supervision should come from an \emph{independent, external} medical vision encoder rather than the model's own visual representations.
To validate this, we compare three choices for the ROI feature source used in $\mathcal{L}_{\text{visual}}$:
\textbf{(a) Qwen3-VL-8B} (self-referential): ROI features are extracted from the backbone's own vision encoder by pooling over mask-aligned patch tokens. This mirrors the design of MedLVR~\citep{xi2026medlvr}, where the model aligns its latent states to its own visual representations.
\textbf{(b) BiomedCLIP}~\citep{zhang2023biomedclip}: an external CLIP-based encoder pretrained on 15M PubMed image-text pairs ($224{\times}224$, $14{\times}14{=}196$ patches, $d_v{=}768$).
\textbf{(c) MedSigLIP}~\citep{sellergren2025medgemma}: an external SigLIP variant medically tuned as part of MedGemma ($448{\times}448$, $32{\times}32{=}1{,}024$ patches, $d_v{=}1{,}152$).
All three variants use the same training data, model architecture, and hyperparameters; only the frozen encoder providing the visual supervision target differs.

\begin{table}[h]
\centering
\small
\caption{\textbf{Ablation on vision encoder for visual supervision.} Self-referential supervision (the model's own encoder) underperforms independent external encoders. Avg.\ accuracy (\%) is reported.}
\label{tab:ablation_encoder}
\setlength{\tabcolsep}{4pt}
\begin{tabular}{@{}llccc@{}}
\toprule
\textbf{Encoder} & \textbf{Type} & \textbf{In-D.} & \textbf{RAD} & \textbf{In-H.} \\
\midrule
(a) Qwen3-VL-8B & Self-referential & 64.73 & 69.18 & 60.90 \\
(b) BiomedCLIP & External & 55.72 & 59.42 & 54.60 \\
\rowcolor{orange!15} (c) MedSigLIP & External & \textbf{81.08} & \textbf{74.28} & \textbf{80.50} \\
\bottomrule
\end{tabular}
\end{table}

\paragraph{Analysis.}
The results reveal a nuanced picture.
MedSigLIP (c) achieves the best performance across all three evaluation sets by a large margin (+16.35 over Qwen3-VL-8B on in-domain, +19.60 on in-house), confirming the value of high-resolution external visual supervision.
Interestingly, the self-referential variant (a) outperforms BiomedCLIP (b) despite providing a theoretically weaker supervision signal.
We attribute this to BiomedCLIP's low spatial resolution ($14{\times}14{=}196$ patches at $224{\times}224$ input): its coarse patch grid produces spatially imprecise ROI features that are too blurry to provide meaningful fine-grained guidance, effectively giving the model a noisy target that harms optimization.
The self-referential approach, while limited by its circular nature, at least operates at the backbone's native resolution and produces coherent patch features.
MedSigLIP combines the best of both worlds: it is an independent external encoder (avoiding the trivial-satisfaction problem of self-referential alignment) with $5.2{\times}$ higher spatial resolution than BiomedCLIP (1{,}024 patches at $448{\times}448$), yielding discriminative and precisely localized ROI features (see qualitative comparison in Figure~\ref{fig:encoder_compare}).
This result validates our design choice and provides direct empirical evidence that (i)~external supervision is superior to self-referential alignment when the external encoder has sufficient spatial resolution, and (ii)~spatial resolution of the supervision source is a critical factor for effective visual grounding in latent reasoning.

\subsection{Ablation on ROI Extraction Strategy}
See~\S\ref{app:roi_grid_search} for hyperparameter grid search results.

\subsection{Ablation on Shared vs.\ Per-Step Visual Target}
\label{app:perstep_visual}

A natural question is whether providing step-specific visual supervision at different granularities could outperform a single shared ROI target.
To investigate this, we design a \emph{per-step visual target} variant that provides progressively finer-grained supervision to each latent step, and compare it against our default shared-target design.

\paragraph{Per-step target construction.}
For each training sample, we construct a sequence of $K{=}4$ visual targets via linear interpolation between a global image feature $\mathbf{f}_{\text{global}}$ and the ROI feature $\mathbf{f}_{\text{ROI}}$:
\begin{equation}
\mathbf{f}_k = \text{Normalize}\!\left(\alpha_k \cdot \mathbf{f}_{\text{ROI}} + (1 - \alpha_k) \cdot \mathbf{f}_{\text{global}}\right),
\end{equation}
where $\alpha_k \in \{0.0, 0.33, 0.67, 1.0\}$ for $k = 1, \dots, 4$.
The global feature $\mathbf{f}_{\text{global}} \in \mathbb{R}^{d_v}$ is computed as the $\ell_2$-normalized mean of all 1{,}024 patch features from the frozen MedSigLIP encoder (without any mask guidance).
This creates a coarse-to-fine curriculum: $z_1$ is supervised toward a holistic image-level representation, while $z_4$ is supervised toward the precise ROI feature.

\paragraph{Training protocol.}
The per-step variant modifies only the visual loss computation: at each step $k$, the visual projector output is regressed against $\mathbf{f}_k$ instead of the shared $\mathbf{f}_{\text{ROI}}$.
All other settings (model architecture, LoRA configuration, semantic supervision, curriculum schedule, and hyperparameters) remain identical to the default VITAL configuration.

\begin{table}[h]
\centering
\small
\caption{\textbf{Ablation on shared vs.\ per-step visual target.} Avg.\ accuracy (\%) is reported.}
\label{tab:ablation_perstep}
\setlength{\tabcolsep}{4pt}
\begin{tabular}{@{}lccc@{}}
\toprule
\textbf{Visual Target Strategy} & \textbf{In-D.} & \textbf{RAD} & \textbf{In-H.} \\
\midrule
\rowcolor{orange!15} (a) Shared (all steps $\to$ $\mathbf{f}_{\text{ROI}}$) & \textbf{81.08} & \textbf{74.28} & \textbf{80.50} \\
(b) Per-step (interpolated $\alpha_k$) & 72.02 & 64.75 & 67.80 \\
\bottomrule
\end{tabular}
\end{table}

\paragraph{Analysis.}
The shared-target strategy consistently outperforms the per-step variant by a substantial margin: +9.06 on in-domain, +9.53 on VQA-RAD, and +12.70 on the held-out in-house testset.
We attribute this to two factors.
First, the per-step design imposes an artificial coarse-to-fine curriculum on the visual pathway, which conflicts with the reasoning trajectory that the model naturally learns under semantic supervision: each $z_k$ reconstructs a distinct reasoning sentence whose content is not necessarily ordered from global to local, so forcing the visual target to follow a fixed interpolation schedule creates gradient conflicts between the two supervision signals.
Second, the shared target provides a consistent gradient direction at every step. Since $\mathcal{L}_{\text{text}}$ already differentiates the steps (each $z_k$ must reconstruct a different reasoning sentence), the shared visual anchor serves as a stable spatial attractor that complements rather than competes with the per-step semantic signal.
The modality collapse analysis (\S\ref{app:modality_collapse}) further confirms that a shared target suffices to produce progressively differentiated latent states without any explicit step-wise visual curriculum.

\subsection{Modality Collapse Analysis}
\label{app:modality_collapse}

To quantify whether latent states degenerate into a homogeneous representation across reasoning steps (i.e., modality collapse), we compute the pairwise cosine similarity between all latent states $\{z_1, \dots, z_K\}$ within each sample and average over the in-domain test set.
Formally, for a given model variant, let $x$ denote a test sample drawn from the in-domain test set $\mathcal{D}_{\text{test}}$, and let $z_i(x) \in \mathbb{R}^d$ be the $i$-th latent state produced by the recurrent loop on input $x$.
We define the inter-step similarity matrix $\mathbf{S} \in \mathbb{R}^{K \times K}$ as:
\begin{equation}
S_{ij} = \mathbb{E}_{x \sim \mathcal{D}_{\text{test}}} \left[ \frac{z_i(x)^T z_j(x)}{\|z_i(x)\| \cdot \|z_j(x)\|} \right].
\end{equation}
A collapsed model produces $S_{ij} \approx 1$ for all $i, j$, meaning that successive reasoning steps encode nearly identical information and the recurrent loop is effectively a no-op.
A healthy model should exhibit lower off-diagonal values, with similarity decreasing as the step distance $|i-j|$ increases, indicating that each step contributes distinct, progressively refined information.

\begin{figure*}[h]
\centering
\includegraphics[width=\textwidth]{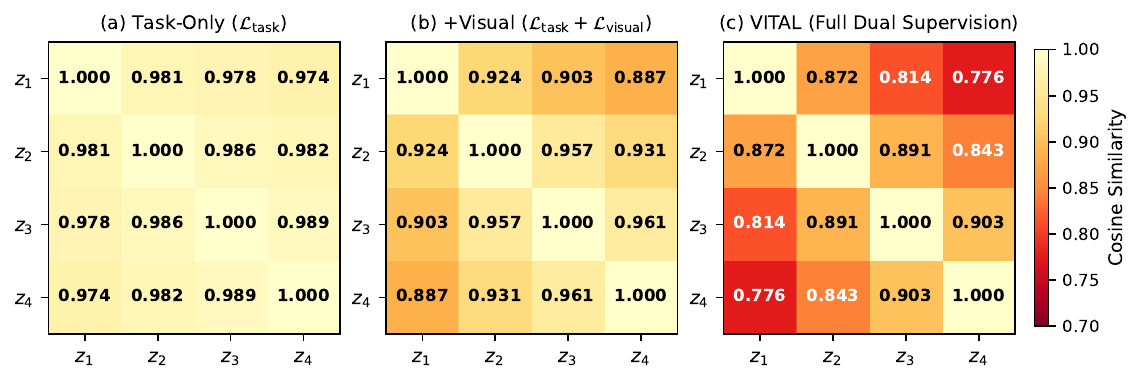}
\caption{\textbf{Inter-step cosine similarity matrices averaged over the in-domain test set.} Lower off-diagonal values (darker cells) indicate more differentiated latent states across reasoning steps. (a) Task-Only exhibits near-total collapse; (c) VITAL maintains healthy inter-step diversity.}
\label{fig:modality_collapse}
\end{figure*}

Figure~\ref{fig:modality_collapse} visualizes $\mathbf{S}$ for three variants: (a)~Task-Only, (b)~+Visual, and (c)~VITAL.
The Task-Only variant shows severe collapse: all off-diagonal entries exceed 0.97, confirming that without auxiliary supervision the latent loop degenerates into near-identical copies of the initial state.
Adding visual supervision alone (b) partially alleviates collapse ($S_{14}$ drops from 0.974 to 0.887), but the middle steps remain highly correlated ($S_{23} = 0.957$), suggesting that a single supervision signal cannot uniformly prevent stagnation across all steps.
With full dual supervision (c), VITAL achieves substantially lower inter-step similarity: $S_{14} = 0.776$ and even adjacent steps maintain meaningful differentiation ($S_{12} = 0.872$, $S_{34} = 0.903$).
This confirms that the combination of semantic and visual supervision forces each latent state to encode genuinely distinct information, with the semantic branch ensuring progressive reasoning refinement and the visual branch anchoring each step to spatial evidence.

\subsection{Quantitative Reliability of the Auxiliary Interpretability Modules}
\label{app:auxiliary_reliability}

The preceding qualitative visualizations show that the auxiliary text decoder and visual projector provide complementary views of VITAL's latent reasoning process.
To quantify the reliability of these post-hoc interpretations, we re-attach both modules and evaluate them on the full in-domain test set, comprising 4{,}607 samples and 9{,}582 valid step-level latent--reasoning pairs.
We evaluate whether the text decoder recovers the semantic content associated with each latent state and whether the visual projector localizes the corresponding image evidence.

\paragraph{Text decoder reconstruction quality.}
For every valid latent state $z_k$, we auto-regressively decode a textual explanation and compare it with the ground-truth reasoning sentence at step $k$.
We report ROUGE-L to measure lexical and sequence-level overlap and BERTScore-F1 to measure semantic similarity.

\begin{table}[h]
\centering
\small
\caption{\textbf{Step-level reconstruction quality of the auxiliary text decoder.} Scores are computed between the decoded explanation from $z_k$ and its ground-truth reasoning sentence. The average is taken over all 9{,}582 valid step-level pairs. All scores are percentages.}
\label{tab:text_decoder_reconstruction}
\setlength{\tabcolsep}{6pt}
\begin{tabular}{@{}cccc@{}}
\toprule
\textbf{Step $k$} & \textbf{Pairs} & \textbf{ROUGE-L $\uparrow$} & \textbf{BERTScore-F1 $\uparrow$} \\
\midrule
1 & 4{,}607 & 54.86 & 86.19 \\
2 & 2{,}880 & 50.93 & 84.32 \\
3 & 1{,}127 & 47.82 & 84.11 \\
4 & 968     & 46.65 & 83.76 \\
\midrule
\textbf{Avg.} & \textbf{9{,}582} & \textbf{52.02} & \textbf{85.14} \\
\bottomrule
\end{tabular}
\end{table}

As shown in Table~\ref{tab:text_decoder_reconstruction}, the decoded explanations retain high semantic similarity at every reasoning depth: BERTScore-F1 remains above 83\%, with an overall score of 85.14\%.
ROUGE-L remains above 46\% at all steps and averages 52.02\%, showing that the decoder also recovers substantial lexical and structural content.
Both metrics decrease moderately with reasoning depth.
Later steps generally synthesize more complex descriptions and spatial relationships, making exact reconstruction more difficult; nevertheless, the relatively stable BERTScore-F1 indicates that semantic content is largely preserved even when the surface wording differs from the reference.
These results quantitatively support the use of the decoder as a textual window into the information encoded by the latent states.

\paragraph{Visual projector localization quality.}
For each latent state $z_k$, we compute the cosine similarity between $\texttt{VP}(z_k)$ and every patch feature in MedSigLIP's $32{\times}32$ feature grid, producing a patch-level activation map.
We compare this map with the ground-truth segmentation mask using three complementary metrics:
(i)~\emph{pointing accuracy}, which measures whether the maximally activated patch lies inside the ground-truth region;
(ii)~\emph{patch IoU}, computed between the top 50\% activated patches and the mask-aligned ground-truth patches; and
(iii)~\emph{center distance}, the normalized Euclidean distance between the activation centroid and the ground-truth mask centroid, for which lower values are better.
We compare VITAL with the Task-Only latent variant to determine whether the localization behavior arises from the auxiliary supervision.

\begin{table}[h]
\centering
\small
\caption{\textbf{Step-level localization accuracy of the visual projector.} Pointing accuracy and patch IoU are percentages; center distance is normalized by image size. The overall results are micro-averaged over all valid step-level samples rather than equally averaging the four step-wise scores, as the number of valid samples differs substantially across reasoning depths. Bold values indicate the better result at each step.}
\label{tab:visual_projector_localization}
\setlength{\tabcolsep}{3.5pt}
\resizebox{\columnwidth}{!}{%
\begin{tabular}{@{}clccc@{}}
\toprule
\textbf{Step $k$} & \textbf{Method} & \textbf{Pointing Acc. $\uparrow$} & \textbf{Patch IoU $\uparrow$} & \textbf{Center Dist. $\downarrow$} \\
\midrule
\multirow{2}{*}{1} & Task-Only & 31.2 & 14.8 & 0.2847 \\
                    & \textbf{VITAL} & \textbf{57.3} & \textbf{28.4} & \textbf{0.1842} \\
\midrule
\multirow{2}{*}{2} & Task-Only & 33.5 & 15.6 & 0.2791 \\
                    & \textbf{VITAL} & \textbf{68.1} & \textbf{37.6} & \textbf{0.1253} \\
\midrule
\multirow{2}{*}{3} & Task-Only & 30.8 & 14.3 & 0.2903 \\
                    & \textbf{VITAL} & \textbf{71.4} & \textbf{44.9} & \textbf{0.0874} \\
\midrule
\multirow{2}{*}{4} & Task-Only & 32.1 & 15.1 & 0.2824 \\
                    & \textbf{VITAL} & \textbf{74.2} & \textbf{51.2} & \textbf{0.0651} \\
\midrule
\multirow{2}{*}{\textbf{Overall}} & Task-Only & 32.3 & 15.2 & 0.2823 \\
                    & \textbf{VITAL} & \textbf{65.4} & \textbf{37.3} & \textbf{0.1334} \\
\bottomrule
\end{tabular}%
}
\end{table}

Table~\ref{tab:visual_projector_localization} shows a clear separation between VITAL and Task-Only latent training.
At the final reasoning step, VITAL achieves 74.2\% pointing accuracy and 51.2\% patch IoU, while reducing normalized center distance to 0.0651.
Moreover, all three localization metrics improve monotonically from $z_1$ to $z_4$: pointing accuracy increases by 16.9 points, patch IoU by 22.8 points, and center distance decreases by 0.1191.
This progression quantitatively corroborates the coarse-to-fine spatial refinement observed in Figures~\ref{fig:qua}, \ref{fig:vp_visualization}, \ref{fig:case1}, and~\ref{fig:case2}.

In contrast, Task-Only produces weak and nearly unchanged localization across reasoning steps, with pointing accuracy remaining between 30.8\% and 33.5\% and patch IoU between 14.3\% and 15.6\%.
The comparison demonstrates that task supervision alone does not yield the progressive grounding exhibited by VITAL, and that the auxiliary supervision is critical for aligning latent reasoning states with spatial evidence.
Together with the text reconstruction results, these measurements provide quantitative evidence that the two re-attached modules recover complementary and meaningful information from VITAL's latent states: semantic reasoning content through the text decoder and localized visual evidence through the visual projector.

\subsection{More Case Studies}
\label{app:case_studies}

To further validate that VITAL's latent states progressively converge toward the target region throughout the reasoning process, we visualize the visual projector output at every latent step for representative samples spanning five imaging modalities: CT, X-ray, pathology, dermoscopy, and ultrasound.
For each sample ($K{=}4$), we project every latent state $z_1, \dots, z_4$ through the trained visual projector and compute patch-level cosine similarity with the resulting feature vector in MedSigLIP's feature space, producing a per-step activation heatmap overlaid on the original image.

\begin{figure*}[t]
\centering
\includegraphics[width=\textwidth]{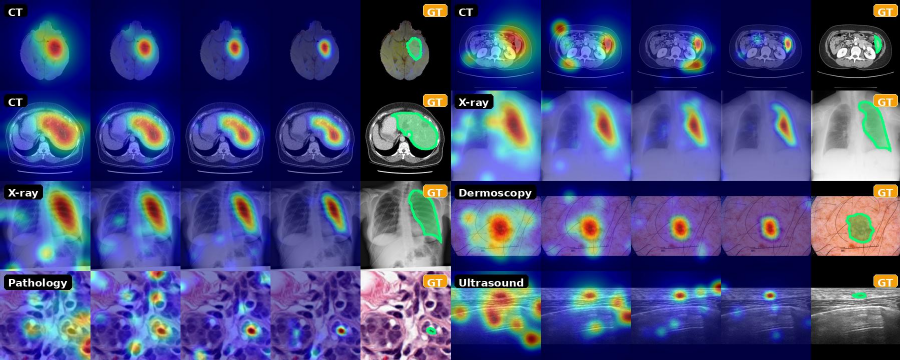}
\caption{\textbf{Progressive visual projector activations across latent reasoning steps ($z_1 \!\to\! z_4$).} The figure shows 8 cases (two per row) spanning CT, X-ray, pathology, dermoscopy, and ultrasound. For each case, the first four images display the patch-level activation heatmap at each latent step, and the fifth image shows the ground-truth region (green overlay). Warmer colors indicate higher cosine similarity with the projected latent feature. Across all modalities, activations progressively sharpen from diffuse early-step responses to precise final-step localization that closely matches the ground truth.}
\label{fig:vp_visualization}
\end{figure*}

Figure~\ref{fig:vp_visualization} presents 8 cases arranged in a $4 \times 10$ grid (two cases per row).
For each case, the first four images show the heatmap evolution from $z_1$ to $z_4$, and the fifth image displays the ground-truth segmentation mask (green overlay) for reference.
Several consistent patterns emerge across modalities:
(i)~\textbf{progressive spatial refinement}: early latent states ($z_1$, $z_2$) produce broad, diffuse activations covering a coarse anatomical neighborhood, while later states ($z_3$, $z_4$) sharpen into tight, well-localized focus on the target structure, mirroring a coarse-to-fine reasoning process;
(ii)~\textbf{convergence to ground truth}: the final-step heatmap ($z_4$) closely aligns with the green GT mask across all modalities, confirming that the recurrent loop successfully drives latent states toward spatially precise visual grounding;
(iii)~\textbf{cross-modality robustness}: the progressive refinement pattern is consistent regardless of imaging modality, target size, or tissue contrast, demonstrating that dual supervision learns a modality-agnostic latent-to-visual mapping.
These qualitative results complement the inter-step similarity analysis (\S\ref{app:modality_collapse}) by showing that the decreasing cosine similarity between successive steps corresponds to meaningful spatial refinement rather than random drift.

We now present two detailed case studies that illustrate VITAL's dual interpretability and compare its reasoning quality against representative baselines.

\begin{figure*}[t]
\centering
\includegraphics[width=\textwidth]{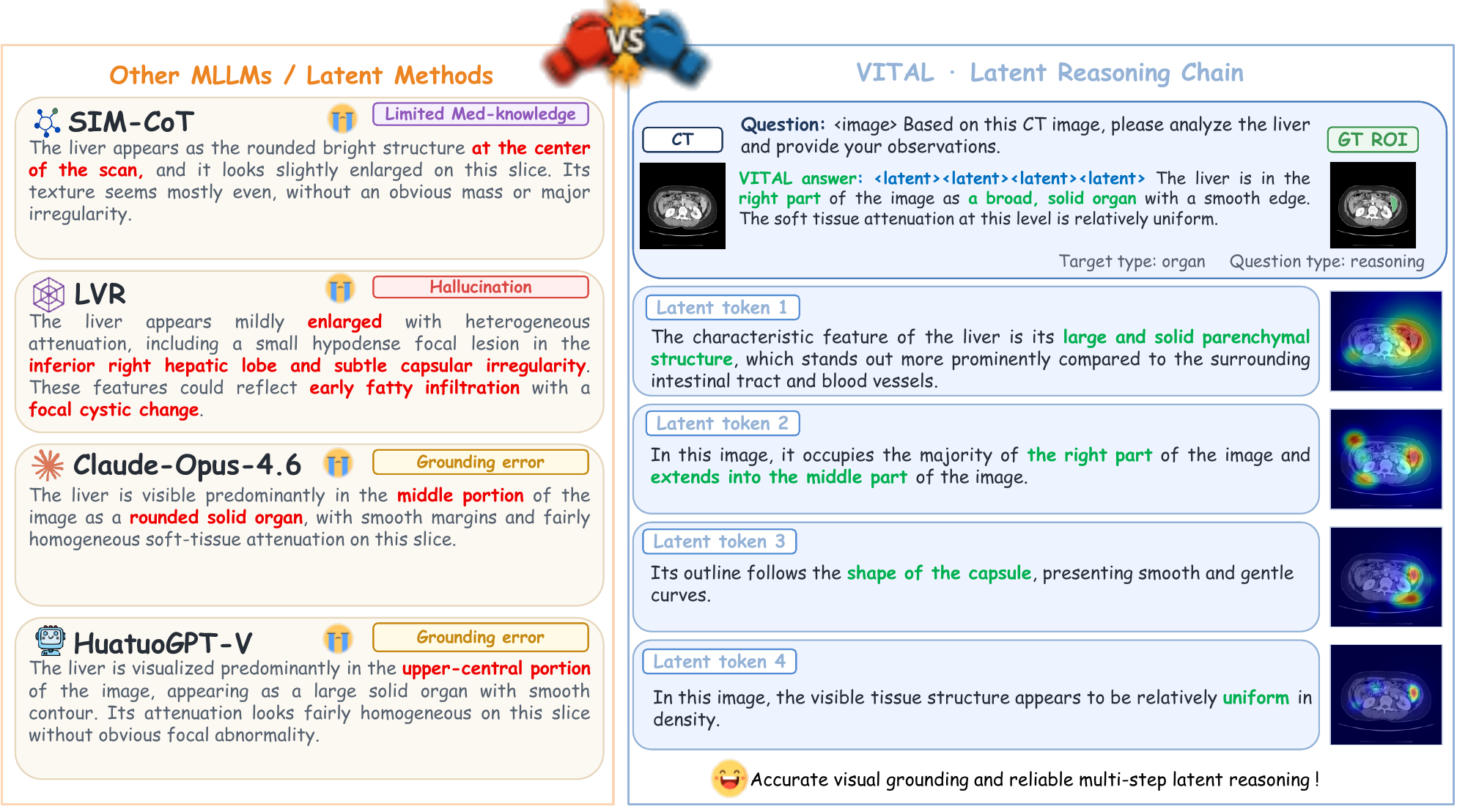}
\caption{\textbf{Case Study 1: CT liver analysis.} \textbf{Right:} VITAL's latent reasoning chain ($z_1 \!\to\! z_4$) with decoded text and visual projector heatmaps showing progressive spatial refinement toward the liver. \textbf{Left:} Baseline outputs exhibiting limited medical knowledge (SIM-CoT), hallucination (LVR), and grounding errors (Claude-Opus-4.6, HuatuoGPT-V).}
\label{fig:case1}
\end{figure*}

\paragraph{Case 1: CT liver analysis (Figure~\ref{fig:case1}).}
The question asks the model to analyze the liver in an abdominal CT image.
VITAL's latent reasoning chain reveals a structured four-step process:
$z_1$ identifies the liver's large, solid parenchymal structure and its contrast against the surrounding intestinal tract and blood vessels;
$z_2$ localizes it to the right part of the image extending into the middle;
$z_3$ characterizes its capsule-like outline with smooth, gentle curves;
$z_4$ assesses the tissue density as relatively uniform.
The accompanying visual projector heatmaps progressively concentrate from a broad abdominal activation to a tight focus on the liver parenchyma, closely matching the GT ROI.
The final answer correctly synthesizes all observations: the liver is in the right part of the image as a broad, solid organ with smooth edges and relatively uniform soft tissue attenuation.

In contrast, the baselines exhibit distinct failure modes.
SIM-CoT suffers from \emph{limited medical knowledge}: it incorrectly describes the liver as located ``at the center of the scan'' and provides only generic observations without precise spatial grounding.
LVR produces outright \emph{hallucinations}, fabricating findings of ``heterogeneous attenuation,'' ``a small hypodense focal lesion in the inferior right hepatic lobe,'' and ``early fatty infiltration with a focal cystic change'' that are entirely absent from the image.
Claude-Opus-4.6 commits a \emph{grounding error} by placing the liver in the ``middle portion'' of the image, and HuatuoGPT-Vision similarly mislocalizes it to the ``upper-central portion.''
These errors highlight the difficulty of accurate spatial reasoning without explicit visual grounding, which VITAL's dual supervision directly addresses.

\begin{figure*}[t]
\centering
\includegraphics[width=\textwidth]{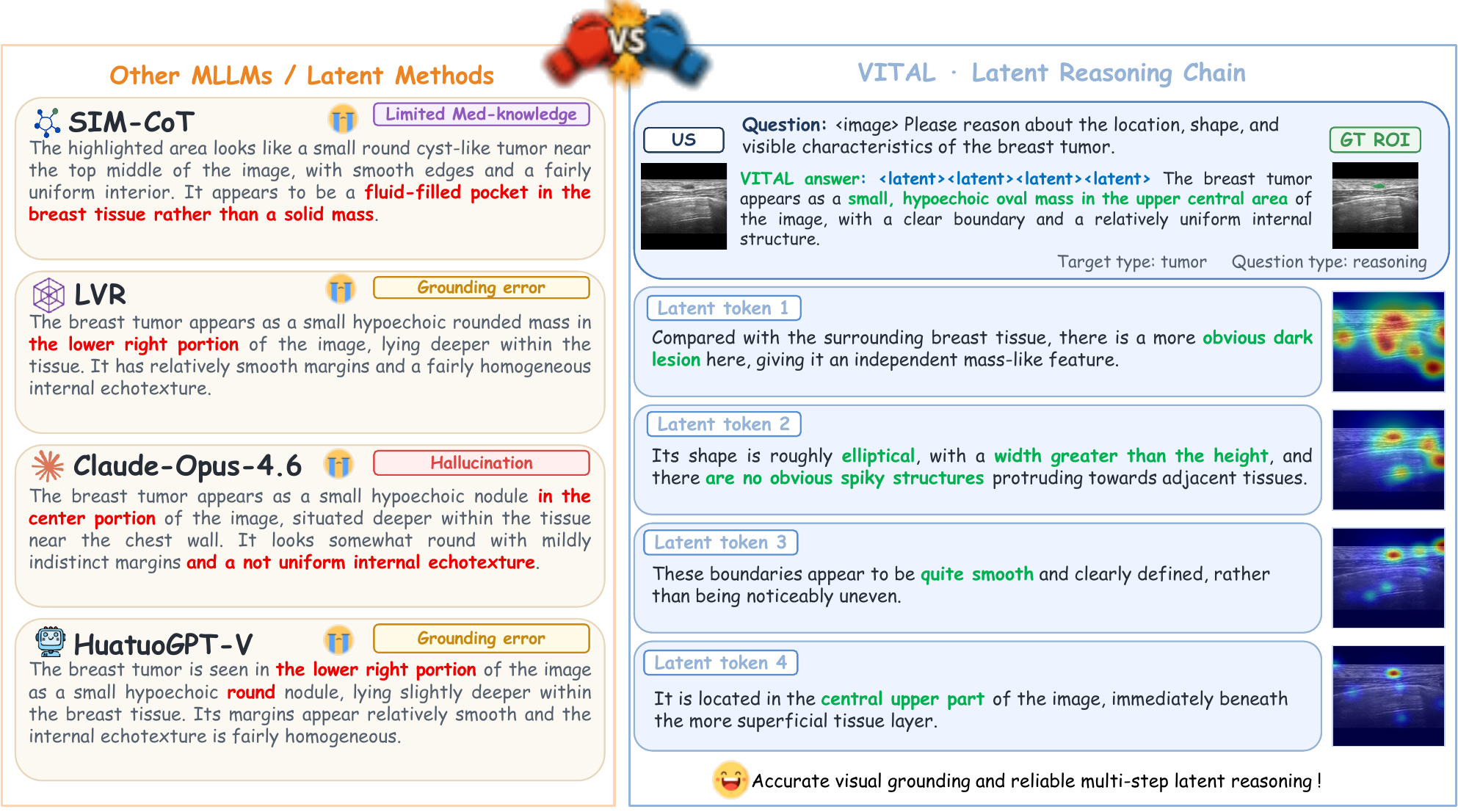}
\caption{\textbf{Case Study 2: Ultrasound breast tumor characterization.} \textbf{Right:} VITAL's latent reasoning chain ($z_1 \!\to\! z_4$) progressively identifies echotexture contrast, elliptical shape, smooth boundaries, and precise localization. \textbf{Left:} Baselines exhibit limited medical knowledge (SIM-CoT), grounding errors (LVR, HuatuoGPT-V), and hallucination (Claude-Opus-4.6).}
\label{fig:case2}
\end{figure*}

\paragraph{Case 2: Ultrasound breast tumor characterization (Figure~\ref{fig:case2}).}
The question asks the model to reason about the location, shape, and visible characteristics of a breast tumor in an ultrasound image.
VITAL's latent chain proceeds as follows:
$z_1$ identifies a region of notably darker echotexture compared to the surrounding breast tissue, recognizing it as a distinct mass-like feature;
$z_2$ characterizes the shape as roughly elliptical with width greater than height, noting the absence of spiky protrusions into adjacent tissues;
$z_3$ assesses the boundary quality as smooth and clearly defined rather than irregular;
$z_4$ localizes the tumor to the central upper part of the image, immediately beneath the superficial tissue layer.
The visual projector heatmaps sharpen from a diffuse response across the breast parenchyma to a precise localization of the hypoechoic nodule, aligning with the GT ROI.
The final answer accurately describes a small, hypoechoic oval mass in the upper central area with a clear boundary and uniform internal structure.

The baselines again demonstrate characteristic failures.
SIM-CoT shows \emph{limited medical knowledge} by misidentifying the tumor as a ``fluid-filled pocket in the breast tissue rather than a solid mass,'' confusing a solid hypoechoic nodule with a cyst.
LVR commits a \emph{grounding error}, placing the mass in ``the lower right portion'' when it is clearly in the upper center.
Claude-Opus-4.6 \emph{hallucinates} characteristics: it describes the nodule as located ``in the center portion'' with ``mildly indistinct margins and a not uniform internal echotexture,'' contradicting the clearly defined boundary and homogeneous interior visible in the image.
HuatuoGPT-Vision similarly mislocalizes the tumor to ``the lower right portion.''
These cases demonstrate that VITAL's visual-semantic dual supervision enables both spatially accurate grounding and clinically precise characterization that baselines consistently fail to achieve.

\section{Future Work}
\label{app:future_work}

We identify several promising directions for extending VITAL.
\textbf{(1) Adaptive reasoning depth.} The current design fixes $K{=}4$ for all samples. Introducing a learned early-exit mechanism (e.g., confidence-based halting or a termination token) would allow the model to allocate fewer steps to simple questions and more to complex clinical reasoning.
\textbf{(2) Broader data coverage.} Our 61K dataset spans 9 modalities from segmentation annotations. Incorporating radiology reports as an additional supervision source could scale training data by orders of magnitude without requiring pixel-level masks for every sample.
\textbf{(3) Reinforcement learning.} VITAL currently relies on teacher distillation, bounding its quality by the teacher's reasoning ability. Outcome-based or process-based RL rewards could push latent reasoning beyond this ceiling, particularly for complex multi-step clinical scenarios.

\end{document}